\documentclass[10pt,twocolumn,letterpaper]{article}

\usepackage{iccv}
\usepackage{times}
\usepackage{epsfig}
\usepackage{graphicx}
\usepackage{amsmath}
\usepackage{amssymb}
\usepackage{textcomp}

\usepackage{subcaption}
\usepackage{lipsum}

\usepackage[breaklinks=true,bookmarks=false]{hyperref}

\iccvfinalcopy % *** Uncomment this line for the final submission

\def\iccvPaperID{10} % *** Enter the ICCV Paper ID here

\usepackage{xcolor}
\definecolor{darkgreen}{rgb}{0,.5,.0}

\newcommand{\bp}{{\bf p}}
\newcommand{\calC}{\mathcal{C}}

\newcommand{\calD}{\mathcal{D}}
\newcommand{\corr}{\text{corr}}

\newcommand{\inliers}{\text{inliers}}
\newcommand{\pose}{{pose}}
\newcommand{\poses}{{poses}}

\usepackage[english]{babel}
\usepackage[utf8]{inputenc}
\usepackage{algorithmicx}
\usepackage{algorithm}
\usepackage[noend]{algpseudocode}

\usepackage{gensymb}
\usepackage{booktabs}
\newcommand{\ra}[1]{\renewcommand{\arraystretch}{#1}}
\begin{document}

%%%%%%%%% TITLE
\title{CorNet: Generic 3D Corners for 6D Pose Estimation of\\New Objects without Retraining}

\author{Giorgia Pitteri$^{1}$ $\quad 
\quad$ Slobodan Ilic$^{2,3}$ $\quad \quad$
Vincent Lepetit$^1$ \\
	\hspace{0em} 	$^1$Laboratoire Bordelais de Recherche Informatique, 
	Universit\'e de Bordeaux, Bordeaux, France\\
	$^2$ Technische Univers\"at M\"unchen, Germany $\quad\quad$ $^3$ Siemens 
	AG, M\"unchen, Germany \\	
	{\hspace{-0em} $^1$$\texttt{\small 
			\{first.lastname\}@u-bordeaux.fr} \quad\quad$
			$^2$$\texttt{\small Slobodan.Ilic@in.tum.de}$}		 
}

%\author{Giorgia Pitteri\\
%LaBRI, Universit\'e de Bordeaux\\
%Bordeaux, France\\
%{\tt\small giorgia.pitteri@u-bordeaux.fr}
%% For a paper whose authors are all at the same institution,
%% omit the following lines up until the closing ``}''.
%% Additional authors and addresses can be added with ``\and'',
%% just like the second author.
%% To save space, use either the email address or home page, not both
%\and 
%Slobodan Ilic\\
%Siemens AG \\
%Munich, Germany\\
%{\tt\small slobodan.ilic@siemens.com}
%\and
%Vincent Lepetit\\
%LaBRI, Universit\'e de Bordeaux\\
%Bordeaux, France\\
%{\tt\small vincent.lepetit@u-bordeaux.fr}
%}

\maketitle
%\thispagestyle{empty}

%%    This  scenario  is
%% particularly  useful in  industrial context,  where  CAD models  of new  objects
%% already exist  but training phases  are cumbersome and the  objects' appearances
%% can change, for example when they become dirty.

%%%%%%%%% ABSTRACT
\begin{abstract}
 We present a novel approach to the  detection and 3D pose estimation of objects
 in  color images.   Its  main contribution  is  that it  does  not require  any
 training  phases  nor data  for  new  objects, while  state-of-the-art  methods
 typically require  hours of training  time and hundreds of  training registered
 images.   Instead, our  method relies  only  on the  objects' geometries.   Our
 method focuses on  objects with prominent corners, which covers  a large number
 of industrial  objects.  We  first learn  to detect  object corners  of various
 shapes in images and  also to predict their 3D poses,  by using training images
 of a small set  of objects.  To detect a new object in  a given image, we first
 identify its corners from its CAD model;  we also detect the corners visible in
 the  image  and predict  their  3D  poses.   We  then introduce  a  RANSAC-like
 algorithm that  robustly and  efficiently detects and  estimates the  object's 3D
 pose by matching its corners on  the CAD model with their detected counterparts
 in the  image.  Because we  also estimate  the 3D poses  of the corners  in the
 image, detecting only 1 or 2 corners  is sufficient to estimate the pose of the
 object, which  makes the approach robust  to occlusions.  We finally  rely on a
 final check that exploits the full 3D geometry of the objects, in case multiple
 objects  have the  same  corner  spatial arrangement.   The  advantages of  our
 approach  make  it particularly  attractive  for  industrial contexts,  and  we
 demonstrate our approach on the challenging T-LESS dataset.
\end{abstract}

%%%%%%%%% BODY TEXT

\section{Introduction}

%% -*- mode: latex; mode: flyspell -*-

3D object detection and pose estimation are of primary importance for tasks such
as robotic  manipulation, virtual and augmented  reality and they have  been the
focus of  intense research  in recent years,  mostly due to  the advent  of Deep
Learning  based approaches  and  the  possibility of  using  large datasets  for
training such methods.  Methods relying on  depth data acquired by depth cameras
are  robust~\cite{Hinterstoisser12,drost2010CVPR}.  Unfortunately,  active depth
sensors are power hungry or sometimes it is not possible to use them.

 \begin{figure}
\begin{center}
    %\resizebox{\linewidth}{!}{
        \begin{tabular}{cc}
            \includegraphics[width=0.4\linewidth]{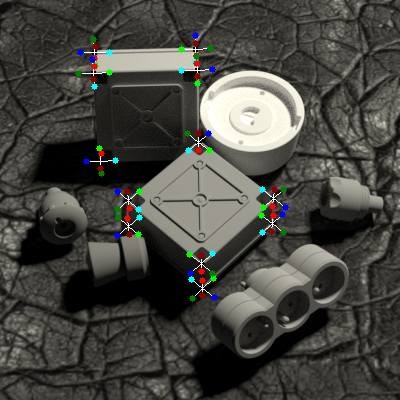} &
            \includegraphics[width=0.4\linewidth]{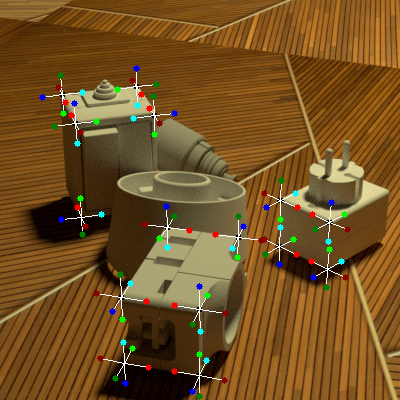} \\
            \includegraphics[width=0.4\linewidth]{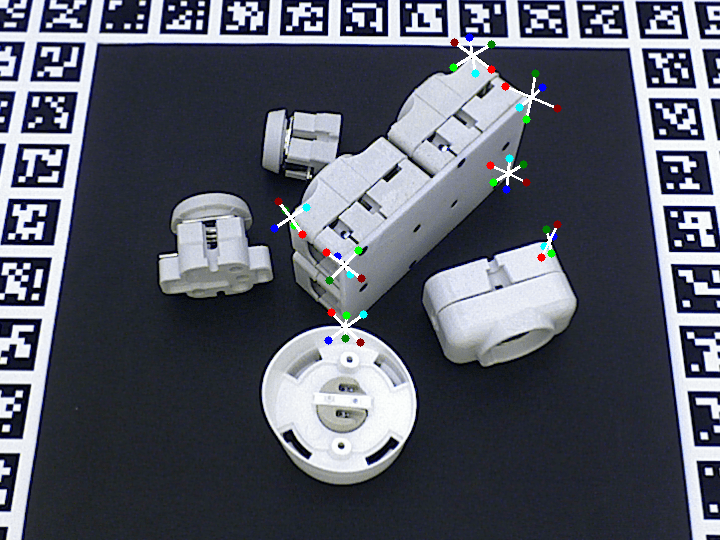} &
            \includegraphics[width=0.4\linewidth]{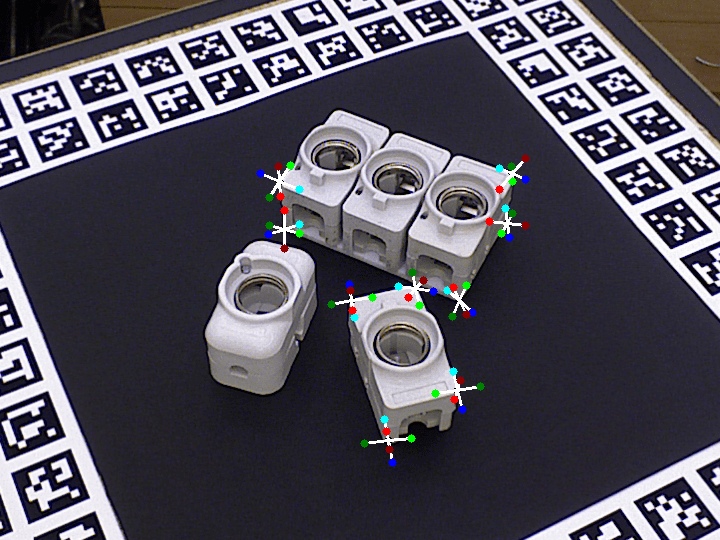}\\
			\includegraphics[width=0.4\linewidth]{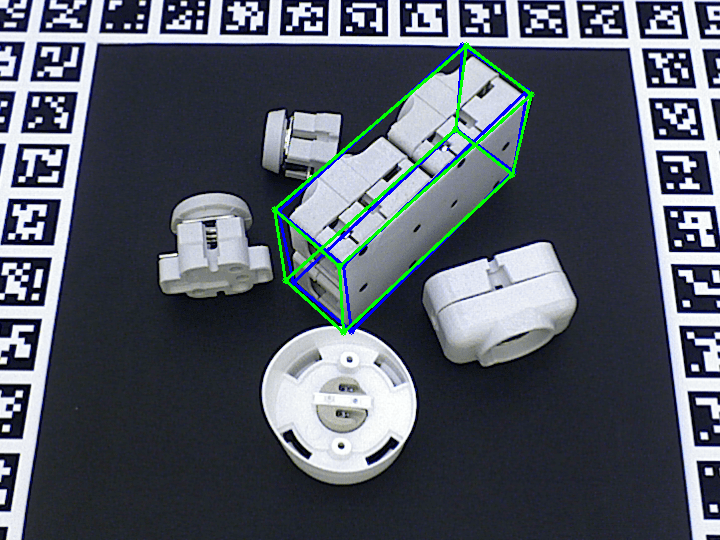} &
            \includegraphics[width=0.4\linewidth]{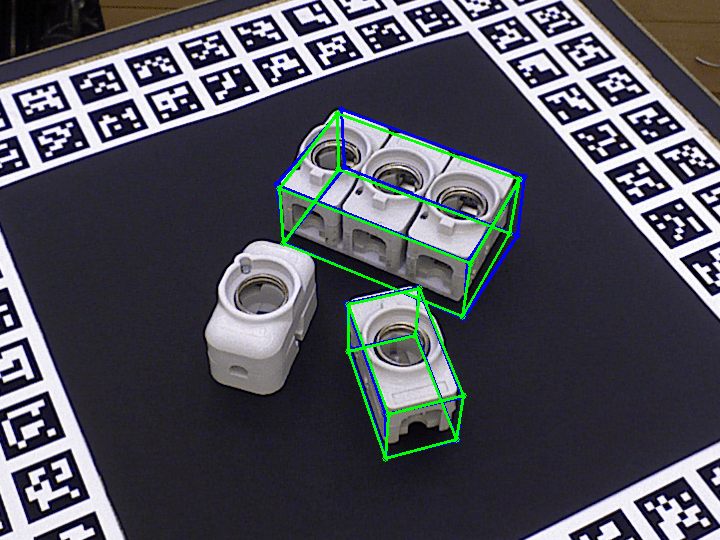}\\
 
        \end{tabular}
    %}
    \caption{Given   a   small  set   of   objects   from  the   T-LESS
        dataset~\cite{Hodan17}, we learn to detect corners of various
        appearances and
        shapes   and   to  estimate   their   3D   poses  using   synthetic
        renderings~(first row).  Then,  given only the CAD model  of new objects
        with corners, we  can detect these objects and estimate  their 3D poses,
        \emph{without any new training phase}~(second and third rows).  The green
        bounding  boxes  correspond to  the  ground  truth  poses and  the  blue
        bounding boxes to the poses estimated with our method. }
\label{fig:intro}
\end{center}
\end{figure}

It is therefore often desirable to rely  on color images, and many methods to do
so         have        been         proposed        recently~\cite{Kehl17,Rad17,
  Tekin2018,Jafari2018,Xiang18,Peng18_PVNet}.   However,  the success  of  these
methods can  be attributed  to supervised Machine  Learning approaches,  and for
each new object, these methods have to  be retrained on many different images of
this object.   Even if domain transfer  methods allow for training  such methods
with                                                                   synthetic
images~\cite{hinterstoisserPreTrainedImageFeatures2017a,Kehl17,Sundermeyer18}
instead                                  of                                 real
ones~\cite{bousmalis17,zhu2017unpaired,ganin2016domain,long2015learning,tzeng2015simultaneous,lee2018diverse,Rad18c,Zakharov2018}
at least  to some  extent, such training  sessions take time,  and it  is highly
desirable to avoid them in practice.

In  this paper,  we propose  a method  that \emph{does  not} require  additional
learning nor training  images for new objects: We consider  a scenario where CAD
models for the target objects exist,  but not necessarily training images.  This
is often the case in industrial settings,  where an object is built from its CAD
model.  We rely  on corners which we  learn to detect and estimate  the 3D poses
during  an  offline   stage.   Our  approach  focuses   on  industrial  objects.
Industrial objects are  often made of similar parts, and  corners are a dominant
common part. Detecting these corners and determining their 3D poses is the basis
for our approach.  We follow a deep  learning approach and train FasterRCNN on a
small set of objects to detect corners and predict their 3D poses.

We use the representation of 3D  poses introduced by \cite{Crivellaro18}: The 3D
pose of  a corner is predicted  in the form of  a set of 2D  reprojections of 3D
virtual points.  This is convenient for  our purpose, since multiple corners can
be easily  combined to compute the  object pose when using  this representation.
However, we need to take care of  a challenge that arises with corners, and that
was ignored in \cite{Crivellaro18}: Because of  its symmetries, the 3D pose of a
corner is  often ambiguous, and  defined only up a  set of rigid  rotations.  We
therefore introduce a robust and efficient algorithm that considers the multiple
possible 3D poses of  the detected corners, to finally estimate  the 3D poses of
the new objects.

In the remainder of the paper, we  review the state-of-the-art on 3D object pose
estimation  from images,  describe our  method, and  evaluate it  on the  T-LESS
dataset, which is made of very challenging objects and sequences.

%-------------------------------------------------------------------------
\section{Related Work}

%% -*- mode: latex; mode: flyspell -*-

In this  section, we first  review recent work on  3D object detection  and pose
estimation from color images.  We also  review works on transfer learning for 3D
pose estimation, as it is a common approach to decrease the number of
real training
images.

\subsection{3D Object Detection and Pose Estimation from Color Images}

Several  recent works  extend  on  deep architectures  developed  for 2D  object
detection by also predicting the 3D  pose of objects.  \cite{Kehl17} trained the
SSD architecture~\cite{Liu16} to  also predict the 3D rotations  of the objects,
and the  depths of the  objects.  To  improve robustness to  partial occlusions,
PoseCNN~\cite{Xiang18}  segments the  objects' masks  and predicts  the objects'
poses  in the  form of  a 3D  translation and  a 3D  rotation. Also  focusing on
occlusion handling,  PVNet~\cite{Peng18_PVNet} proposed a network  that for each
pixel  regresses an  offset  to  predefined keypoints.   Deep-6DPose~\cite{Do18}
relies    on   Mask-RCNN~\cite{He17}.     Yolo3D~\cite{Tekin2018}   relies    on
Yolo~\cite{Redmon16}  and predicts  the  object  poses in  the  form  of the  2D
projections of the  corners of the 3D  bounding boxes, instead of  a 2D bounding
box.   \cite{Rad17} also  used this  representation to  predict the  3D pose  of
objects, and shows how to deal with  some of the ambiguities of the objects from
T-Less---however   it    does   not   provide   a    general   solution.    Some
methods~\cite{Kehl17,hinterstoisserPreTrainedImageFeatures2017a,Sundermeyer18}
use synthetic training images generated from CAD models, but for each new model,
they need to retrain their network, or a new one.

Somewhat             related             to            our             approach,
\cite{Jafari2018,Brachmann16,zakharov2019dpod,Peng18_PVNet} first predict the 3D
coordinates  of  the  image  locations  lying on  the  objects,  in  the  object
coordinate system,  and predict the  3D object pose through  hypotheses sampling
with  preemptive  RANSAC.   Instead  of  predicting the  3D  coordinates  of  2D
locations, \cite{Crivellaro18} predicts the 2D  projections of 3D virtual points
attached to  object parts. The advantage  of this approach is  its robustness to
partial occlusions,  as it  is based on  parts, and the  fact that  the detected
parts can be used  easily together to compute the 3D pose  of the target object.
In this paper, we  rely on a similar representation of parts,  but extends it to
deal with ambiguities, and show how to  use it to detect unknown objects without
retraining.

All these  works require extensive training  sessions for new objects,  which is
what we avoid in  our approach. Previous works, based on  templates, also aim at
avoiding such training sessions. For example, \cite{Hinterstoisser12} proposes a
descriptor  for object  templates, based  on  image and  depth gradients.   Deep
Learning  has also  been applied  to  such approach,  by learning  to compute  a
descriptor       from       pairs        or       triplets       of       object
images~\cite{Wohlhart15,Balntas17,zakharov2017iros,bui2018icra}. Like ours, these
approaches  do not  require  re-training, as  it only  requires  to compute  the
descriptors for images of the new objects. However, it requires many images from
points of view sampled  around the object.  It may be  possible to use synthetic
images,  but then,  some  domain transfer  has  to be  performed.  But the  main
drawback of  this approach is the  lack of robustness to  partial occlusions, as
the descriptor is computed for whole images of objects. It is also not clear how
it would  handle ambiguities, as  it is based on  metric learning on  images. In
fact,  such approach  has been  demonstrated on  the LineMod,  which is  made of
relatively simple objects,  and never on the T-Less dataset,  which is much more
challenging.

%%%%%%%%%%%%%%%%%%%%%%%%%%%%%%%%%%%%%%%%%%%%%%%%%%%%%%%%%%%%%%%%%%%%%%%%%%%%%%%%

\subsection{Transfer Learning for 3D Pose Estimation}

Another approach  to limit  the number  of real  training images  is to  rely on
synthetic images,  which can  be rendered when  a CAD model  is available  as we
assume here.   This is a very  popular approach, which requires  domain transfer
between  synthetic  and  real  images.   Domain  transfer  between  images  from
different  datasets  is a  common  problem  in computer  vision~\cite{Gupta2016,
  Rozantsev17,                                                          Cai2018,
  bousmalis17,zhu2017unpaired,ganin2016domain,long2015learning,tzeng2015simultaneous,lee2018diverse,Rad18c,Zakharov2018},
and we focus here only on works related to 3D pose estimation.

Generative  Adversarial Networks~(GANs)~\cite{Goodfellow14}  have  been used  to
generate              training             images~\cite{Bousmalis16,Mueller2018,
  bousmalis17,zhu2017unpaired,ganin2016domain,long2015learning,tzeng2015simultaneous,lee2018diverse,Rad18c,Zakharov2018},
by making  synthetic images closer to  real images.  However, they  need to have
access  to  target  domain data  and  usually  overfit  to  them and  would  not
generalize well to new domains.   Differently, \cite{Zakharov2018} chose to make
real depth images  closer to clean synthetic depth images.   It requires however
careful  augmentation  to  create   realistic  synthetic  depth  maps.   Because
synthetic  depth maps  are easier  to  render than  color images,  \cite{Rad18c}
proposes to  learn a mapping  between features for  depth maps and  features for
color  images using  an RGB-D  camera.  Another  interesting approach  is domain
randomization~\cite{Tobin2017}, which  generates synthetic training  images with
random appearance, by  varying the object textures and  rendering parameters, to
improve  generalization.    AAE~\cite{Sundermeyer18}  presents   another  domain
randomization approach  based on autoencoders  to train pose  estimation network
from CAD models.

Even if  these works  allow to  reduce the  number of  real images  required for
training,  or to  completely get  rid  of them,  they still  require a  training
session for new objects, which is what we avoid entirely with our approach.

%-------------------------------------------------------------------------
  
\section{Approach}

%% -*- mode: latex; mode: flyspell -*-

We describe  our approach in  this section.  We first  describe how we  learn to
detect corners  and predict  their 3D  poses. We then  present our  algorithm to
estimate  the 3D  poses of  new  objects in  an  input image,  from the  corners
detected in this image.
 
\begin{figure}
  \centering
  \includegraphics[scale=0.25]{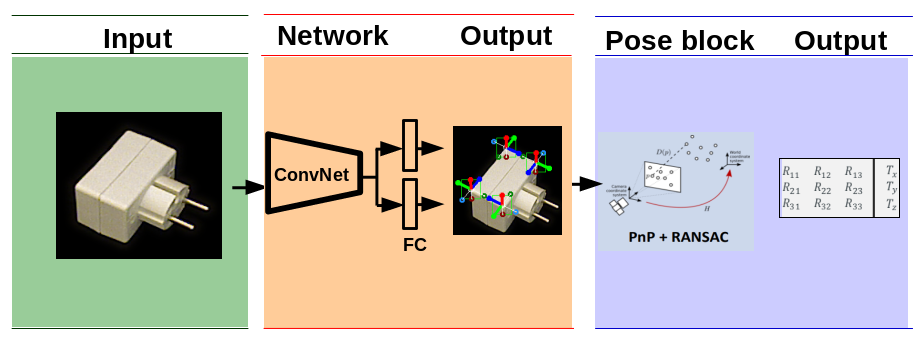}
  \caption{Overview of our approach. We  modified FasterRCNN to detect generic
    corners  in  images  and  predict  their 3D  poses.  Our  pose  estimation
    algorithm, which is an extension of RANSAC, estimates the 3D poses of full
    objects from these detections. }
  \label{fig:points}
\end{figure}

\subsection{Corner Detection and 3D Pose Estimation} \label{parts_detection}
\begin{figure}
  \begin{center}
    \begin{tabular}{cc}
      \includegraphics[width=0.5\linewidth]{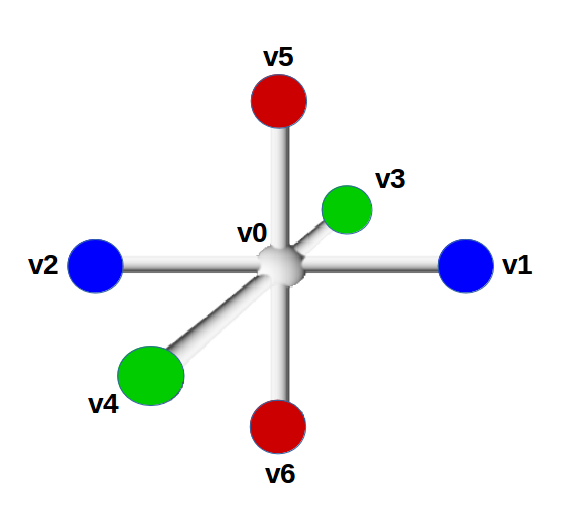} &
      \includegraphics[width=0.4\linewidth]{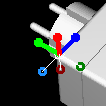} \\
      (a) & (b)\\
    \end{tabular}
  \end{center}
  \begin{center}
    \begin{tabular}{cccc}
      \includegraphics[width=0.2\linewidth]{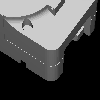} &
      \includegraphics[width=0.2\linewidth]{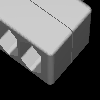} &
      \includegraphics[width=0.2\linewidth]{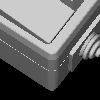} &
      \includegraphics[width=0.2\linewidth]{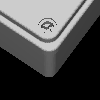} \\
    \end{tabular}
    (c)
  \end{center}
  \vspace{-0.5cm}
  \caption{ \label{fig:pose_representation} 3D pose  representation of an object
    part from~\cite{Crivellaro18}.  (a) Seven 3D control points arranged to span
    3 orthogonal directions are assigned to each part.  (b) Given an image patch
    of  the part,  \cite{Crivellaro18} predicts  the 2D  reprojections of  these
    control points,  and computes  the 3D  pose of the  objects from  these 3D-2D
    correspondences.  The first difference  with \cite{Crivellaro18} is that our
    corners are \emph{generic} in the sense  that they can correspond to corners
    of various  shapes and  appearances, as corners  from different  objects can
    actually be  different (c),  while \cite{Crivellaro18} considers  parts from
    object  instances.   This   allows  us  to  consider   new  objects  without
    retraining.  The  other difference  is  that  we  need  to handle  the  pose
    ambiguities of corners  (due to their symmetries), which  was not considered
    in \cite{Crivellaro18}.  This is done in our 3D pose estimation algorithm. }
\end{figure}

We   use  the   representation   of   the  3D   pose   of   a  part   introduced
in~\cite{Crivellaro18}  to  represent   the  3D  pose  of   our  corners.   This
representation is made  of the 2D reprojections  of a set of  3D control points.
Its main advantage is that it is easy  to combine the 3D poses of multiple parts
to   compute    the   3D   pose   of    the   object   by   solving    a   P$n$P
problem~\cite{Hartley00}.   These control  points are  only ``virtual'',  in the
sense they do  not have to correspond  to specific image features.   As shown in
Fig.~\ref{fig:pose_representation},  we consider  seven 3D  control points  for
each part, arranged to span 3 orthogonal  directions and the center of the part,
as in~\cite{Crivellaro18}.

While  \cite{Crivellaro18}  performed detection  and  pose  prediction with  two
separate networks,  we rely on  the Faster-RCNN framework~\cite{Ren15} as  it is
common practice now  for various problems: We kept the  original architecture to
obtain region proposals that correspond to  parts and added a specific branch to
predict the 2D coordinates of each  control point. This branch is implemented as
a     fully     connected     two-layer     perceptron.  The size  of its
output is  $2\times N_v$,  where $N_v$  is the  number of  control points  for a
detected corner,  and with  $N_v = 7$  in practice.  For  training, we  used the
default  hyperparameters used  in \cite{Ren15}  and  the same  loss function  to
predict the  object class (corner vs  background).  We also added  to the global
loss term, a  squared loss for training the predictions  of the reprojections of
the control points.

To train  FasterRCNN, we  used a  small number  of objects  exhibiting different
types of  corners, shown  in Fig.~\ref{fig:pose_representation}(c),  and created
synthetic images of these objects for training. Two examples of these images are
shown on  the first row  of Fig.~\ref{fig:intro}.   These images are  created by
randomly placing the training objects in a simple scene made of a plane randomly
textured, and randomly lighted.  In practice, we noticed that we did not need to
apply transfer  learning to take  care of the  domain gap between  our synthetic
images and  the real test images  of T-LESS.  This  is probably due to  the fact
that we consider only local parts of  the images, and because the test images of
T-LESS are relatively noise-free.  Given the  CAD models of these objects we can
select the control points in 3D and  project them with the ground truth pose. In
this way,  we obtain the  2D ground truths  reprojections of the  control points
needed to train the network.

\subsection{Ambiguities between Corner Poses and How to Handle Them}
\label{sec:ambiguities}

\begin{figure}
  \begin{center}
    \begin{tabular}{cc}
      \includegraphics[width=0.4\linewidth]{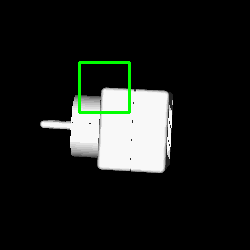} &
      \includegraphics[width=0.4\linewidth]{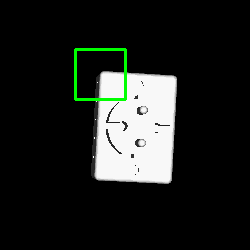} \\
    \end{tabular}
    \caption{The  same corner  can  look the  same under  different  3D poses.  This
      implies that it  is possible to predict the  3D pose of a corner  only up to
      some rigid motions. }
    \label{fig:ambiguities}
  \end{center}
\end{figure}

As shown in  Fig.~\ref{fig:ambiguities}, many ambiguities may  happen when trying
to predict the 3D pose of a corner from its appearance. These ambiguities do not
happen in  the problems considered  by \cite{Crivellaro18},  and are due  to the
symmetries of corners.   If we ignore these ambiguities, we  would consider only
one pose  among all  the possible  poses for each  detected corner,  which would
result in missing new objects very often.

\begin{figure}
  \begin{center}
    %\resizebox{\columnwidth}{!}{
      \begin{tabular}{ccc}
        \includegraphics[width=0.25\linewidth]{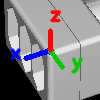} &
        \includegraphics[width=0.25\linewidth]{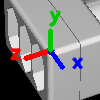} &             
        \includegraphics[width=0.25\linewidth]{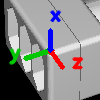} \\
      \end{tabular}
    %}
  \end{center}
  \caption{\label{fig:three_possible_poses} Given the image of a  corner, three arrangements of 3D virtual points are possible.}
\end{figure}

From  the image  of a  corner, there  are  in general  3 possible  3D poses  that
correspond to this image, as shown in Fig.~\ref{fig:three_possible_poses}. Given
one possible 3D  pose $\bp$, it is  possible to generate the two  other poses by
applying rotations around the corner.  In  our case, since we represent the pose
with  the 2D  reprojections of  the virtual  points, this  can also  be done  by
permuting properly the  2D reprojections. We therefore  introduce 2 permutations
$\Sigma_1$ and $\Sigma_2$  which operate on the 2D reprojections  as depicted in
Fig.~\ref{fig:three_possible_poses}.  Given  a pose predicted by  FasterRCNN, we
can generate the  2 other possible poses by applying  $\Sigma_1$ and $\Sigma_2$.
This is used in our pose estimation algorithm described in the next subsection.

\subsection{Pose Estimation Algorithm}
\label{ransac_explanation}

We  represent a  new  object to  detect  as a  set  $\calC =  \{  \calC_1, ..  ,
\calC_{N_C}\}$ of $N_C$  3D corners.  This can  be done using only the  CAD model of
the object. Each corner is made of $N_v$ 3D virtual points: $\calC_i = \{ M_{i,1},
.., M_{i,N_v} \}$ expressed in the object coordinate system.

From our FasterRCNN framework, given an input image, we obtain a set $\calD = \{
d_1, \ldots, d_{N_d} \}$ of $N_D$  detected corners $d_j$.  Each detected corner
$d_j$  is   made  of  $N_v$   predicted  2D  reprojections:  $d_j   =  [m_{j,1},
  \ldots,m_{j,N_v}]$.

The  pseudocode for  our detection  and pose  estimation algorithm  is given  as
Alg.~\ref{alg:pose_estimation_algo}.  To deal with the erroneous detected parts,
we use the same strategy as RANSAC.  By matching the detected corners $d_j$ with
their 3D counterparts  $\calC_i$, it is possible  to compute the 3D  pose of the
object  using a  PnP algorithm.   Since each  corner is  represented by  $N_v=7$
points, it is possible to compute the  pose from a single match. As explained in
Section~\ref{sec:ambiguities}, each detected corner can correspond to 3 possible
arrangements of  virtual points, and we  apply $\Sigma_1$ and $\Sigma_2$  to the
$m_{j,k}$  reprojections to  generate the  3D  possible poses  for the  detected
corners.

In order to find  the best pose among all these 3D possible  poses, we compute a
similarity score as the cross-correlation between the gradients of the image and
the image gradients of the CAD model rendered under the 3D pose. We finally keep
the pose with the largest similarity score as the estimated pose.
 
\begin{algorithm}
  %\algsetup{linenosize=\tiny}
  %\small
  \begin{algorithmic}[1]
    %% \footnotesize
    \State $\calC \gets \{C_i\}_i$, the set of 3D corners on the new
    object. Each 3D corner $C_i$ is made of 7 3D control points, expressed in the
    coordinate system of the new object.
    \State $\calD \gets \{d_j \}_j$, the set of 2D detected corners in the input
    image. Each 2D corner $d_j$ is made of 7 2D image locations.
    \State 
    \Procedure{Pose\_Estimation}{$\calC$, $\calD$}
    \State $\poses \gets []$  \Comment{\small{Set of possible poses and their scores}}
    \For{$C \in \calC$}
    \For{$d \in \calD$}
    \For{$\Sigma \in \{I, \Sigma_1, \Sigma_2\}$}
    \State $\corr \gets (C, \Sigma(d))$ \Comment{\tiny{2D-3D correspondence}}
    \footnotesize
    \State $\pose \gets \textsc{PnP}(\corr)$ \Comment{\tiny{3D pose estimate}}
    \footnotesize
    \State $nb_\inliers \gets \textsc{Compute\_Inliers}(\pose, \calC, \calD)$
    \If{$nb_\inliers > \tau_\inliers$}
    \State $\textsc{Refine}(\pose, \calC, \calD)$ \Comment{\tiny{Compute pose using all the inliers}}
    \footnotesize
    \State $s_\pose \gets \textsc{Score}(\pose, \calC, \calD)$          
    \State Add $(\pose,s_\pose)$ to $\poses$ 
    \EndIf
    \EndFor % end for m
    \EndFor % end for i
    \EndFor % end for j
    \State \textbf{return}  $\pose$ with best $s_\pose$ in $\poses$
    \EndProcedure
    \State 
    \Procedure{Score}{$\pose$, $\calC$, $\calD$}
    \State $s \gets 0$
      \State $template \gets ImageGradients(rendering(model, \pose)$
      \State $edges_{input} \gets ImageGradients(input_{image})$	
      \State $s \gets Cross\_Correlation(edges_{input}, template)$
    \State \textbf{return} $s$
    \EndProcedure
  \end{algorithmic}
  \caption{Pose estimation algorithm}
  \label{alg:pose_estimation_algo}
\end{algorithm}

%-------------------------------------------------------------------------
  
\section{Evaluation}

%% -*- mode: latex; mode: flyspell -*-

In  this section,  we present  and discuss  the results  of our  pose estimation
algorithm.  We  first describe the  metrics used in  the literature and  in this
paper.  Then, we  show  a quantitative  analysis of  object  detection and  pose
estimation as well as qualitative results.   All the results are computed on the
challenging T-LESS dataset~\cite{Hodan17}.

\subsection{Metrics}

To evaluate our  method, we use the percentage of  correctly predicted poses for
each sequence  and each object of  interest, where a pose  is considered correct
based on  the ADD metric.  This  metric is based  on the average distance  in 3D
between the model points after applying  the ground truth pose and the estimated
one. A  pose is  considered correct  if the distance  is less  than 10\%  of the
object's diameter.

\subsection{Results}

The complexity  of the  test scenes  varies from several  isolated objects  on a
clean background  to very  challenging ones with  multiple instances  of several
objects with a high amount of  occlusions and clutters.  Only few previous works
present results on the challenging T-LESS dataset~\cite{Hodan17}. To the best of
our knowledge, the problem of pose estimation  of new objects that have not been
seen at training time has not been addressed yet.

In order to evaluate our method,  we split the objects from T-LESS into two
sets: One set of objects seen by the network during the training and one set of
objects never seen and used for evaluation at testing time. More specifically,
we train our network on corners extracted from Objects \#6, \#19, \#25, \#27
and \#28 and test it on Objects \#7, \#8, \#20, \#26 and \#29 on T-LESS
test scenes \#02, \#03, \#04, \#06, \#08, \#10, \#11, \#12, \#13, \#14 and \#15.

\begin{table}
  \small
  \ra{1.1}
  \resizebox{\columnwidth}{!}{
    \begin{tabular}{@{}llll|l@{}}\toprule
      Scene: Obj & $AD\{D|I\}_{10\%}$ & $AD\{D|I\}_{20\%}$ & $AD\{D|I\}_{30\%}$  & detection [$\%$]\\ \toprule
      02: 7 &   68.3    & 80.1  &  83.7   &  67.3 \\ 
      03: 8 &    57.9  &  72.5  &  78.7  & 76.3 \\
      04: 26 &    28.1 &  47.2  &  56.2  & 48.3 \\
      04: 8 &    21.2 &  53.0  &  68.2  & 35.7\\        
      06: 7 &    36.8 &  61.7  &  78.7  & 73.7 \\
      08: 20 &    10.0  &  40.4  &  56.1  & 34.1 \\ 
      10: 20 &    27.8  &  47.2  &  58.3  & 30.0 \\
      11: 8 &   58.8  &  74.9  &  85.3  & 74.3 \\  
      12: 7 & 23.1  &  44.6 & 47.7 & 54.6 \\ 
      13: 20 &  26.6   & 57.3  & 69.0   &  52.9 \\  
      %15: 26 &  22.0  & 48.6  & 57.8 &  22.0 \\  % don't show results on obj 26
      15: 29 &   48.0  &   59.1 & 76.7 &  38.3 \\ 
      14: 20 & 10.0 & 24.6 & 31.6 & 44.0 \\
      \midrule
          {\bf Average} &  34.7($\pm$18.5)    & 55.2($\pm$15.2)  & 65.9($\pm$15.6) & 52.5($\pm$16.2) \\ 
          \bottomrule
    \end{tabular}
  }
  \caption{\label{tab:results} Our quantitative results on T-Less test Scenes \#02, \#03, \#04, \#06, \#08, \#10, \#11, \#12, \#13, \#14, \#15. Last column reports the detection accuracy.  We consider the object to be detected if the $IoU$ between the rendering of the object with the pose estimate and with the ground truth is higher then $0.4$.}
\end{table}

\subsubsection{2D Detection}

We  first evaluate  our method  in  terms of  2D  detection. Even  this task  is
challenging on  the T-LESS dataset  given our setting,  as the objects  are very
similar to each others.

Most of previous works separate the detection task from the pose estimation. For example, in \cite{Rad17}, the authors present a method that first detects objects through a segmentation approach and then use the corresponding crop of the image to estimate the pose.  Some works only focus on pose estimation~\cite{Sundermeyer18}, and use the ground-truth crops of each objects of the scene to avoid the detection step.

In this work, we cannot access images of objects on which the pose estimation is done. Thus, it is not possible to train a separated object detection network or segmentation network to solve this problem. Our method returns directly the 3D poses of the objects. To evaluate the detection accuracy, we therefore use the 2D bounding boxes computed from the reprojections of the CAD models under the estimated 3D pose.

We report our detection accuracy in the last column of Table~\ref{tab:results}.  The accuracy is measured in terms of \textit{Intersection over Union}~($IoU$) between the rendering of the object with the estimated pose and the rendering of the object with the ground truth pose.  An object is considered correctly detected in the frame if $IoU > 0.4$. Our method succeeds an average of $52.5 \%$  of good detection without any detection or segmentation priors.
 
\subsubsection{3D Pose Estimation}
We evaluate the pose estimation on images  where the object of interest has been
detected.  For each  object  of  our experiments,  we  compute  the ADD  metric
presented   above.  Table~\ref{tab:results}   reports  the   scores  for   three
percentages  of object  diameters. For  symmetrical objects,  we report  the ADI
metric instead  of ADD.   The object  3D orientation  and translation  along the
x-and y-axes  are typically  well estimated.  Most of  the translation  error is
along the  z-axis, as it  is usually  the case of  other algorithms for  3D pose
estimation from color images.

\newcommand{\imsize}{0.18\linewidth}

\begin{figure*}[h]
  \begin{center}
    \begin{tabular}{ccccc}
      \includegraphics[width=\imsize]{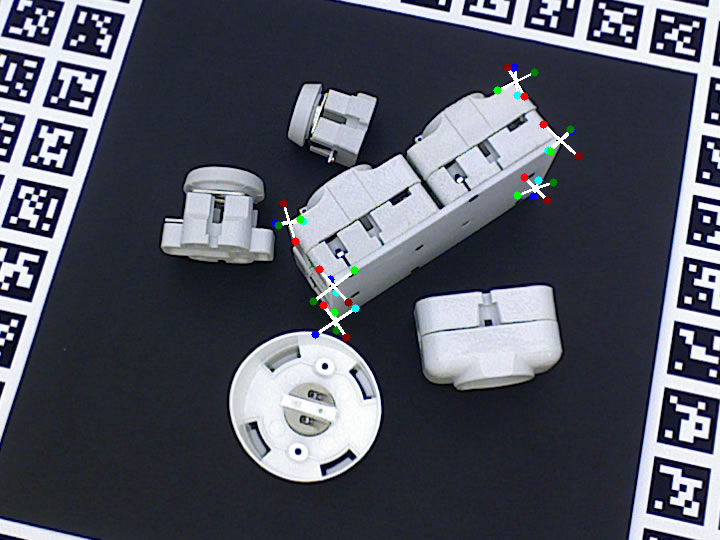} &
      \includegraphics[width=\imsize]{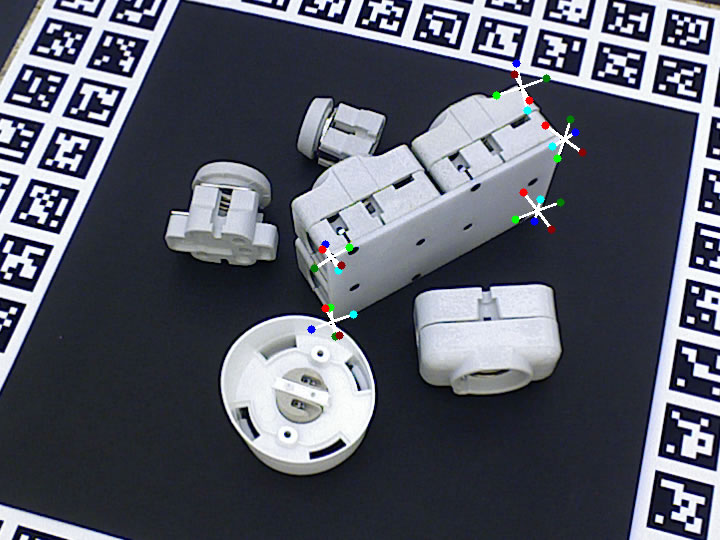} &
      \includegraphics[width=\imsize]{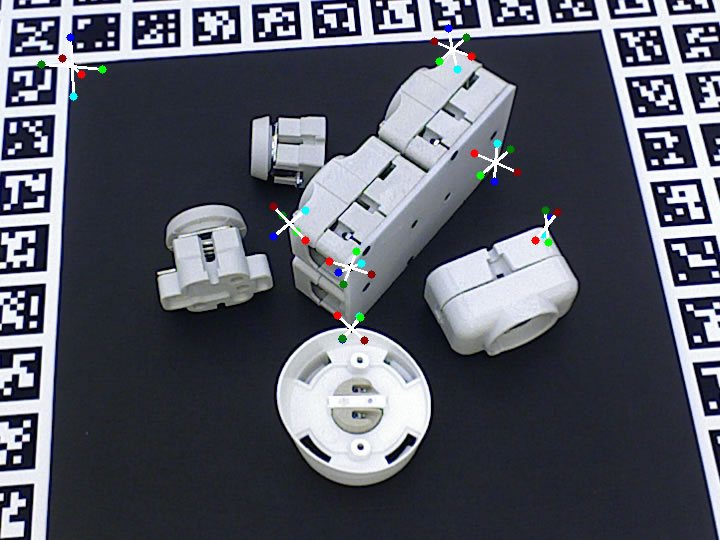} &
      \includegraphics[width=\imsize]{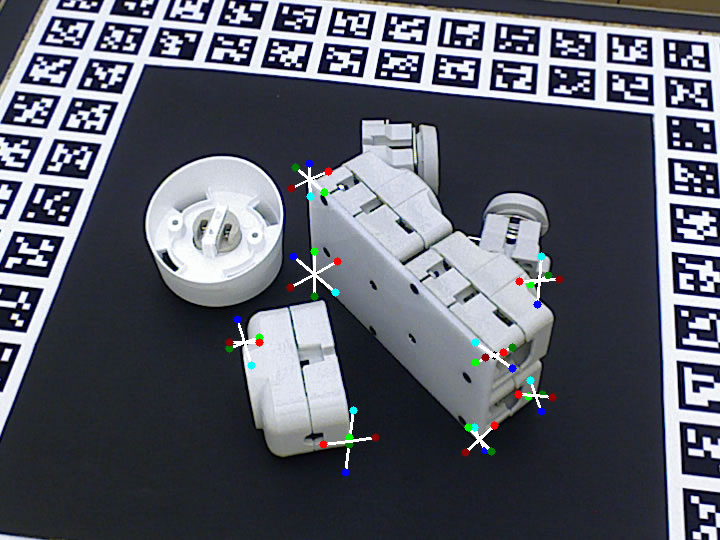} &
      \includegraphics[width=\imsize]{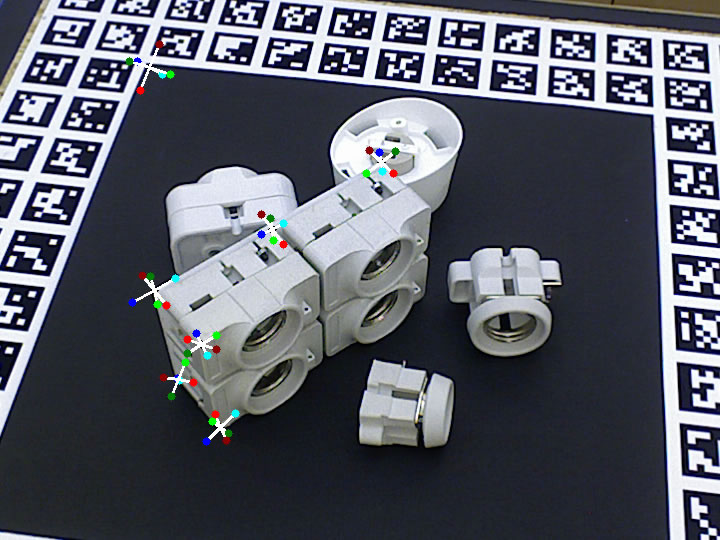} \\
      \includegraphics[width=\imsize]{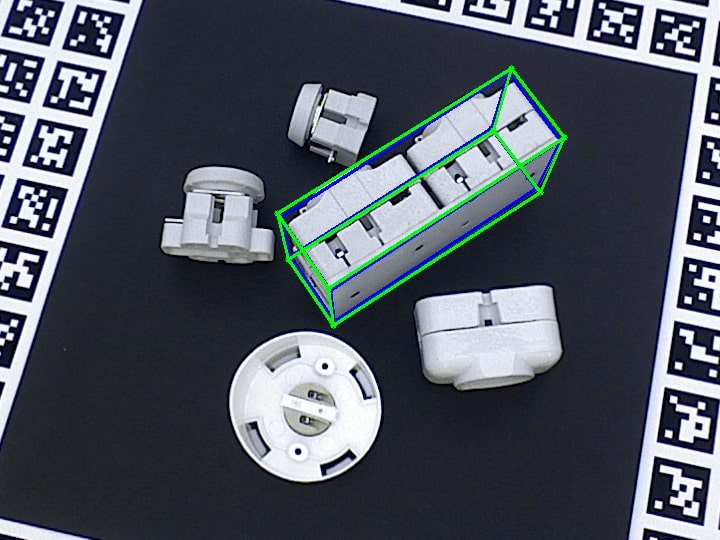} &
      \includegraphics[width=\imsize]{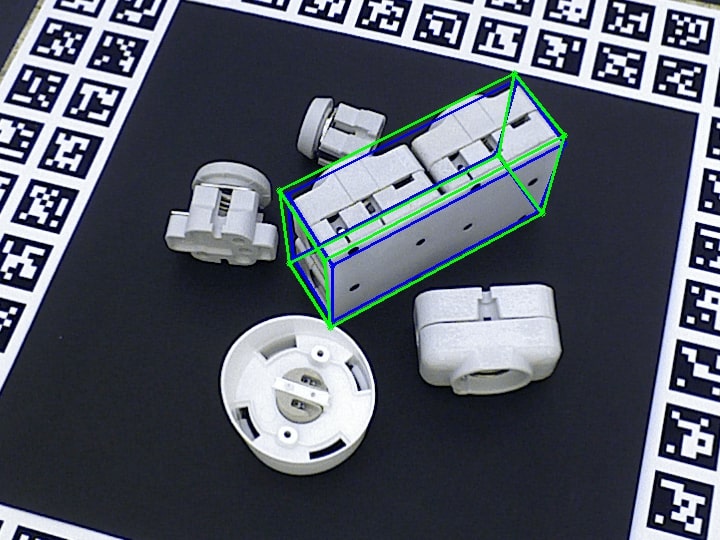} &
      \includegraphics[width=\imsize]{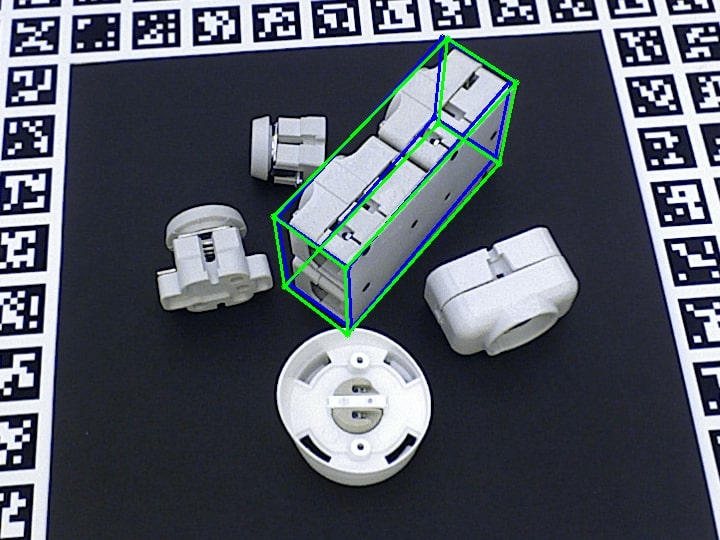} &
      \includegraphics[width=\imsize]{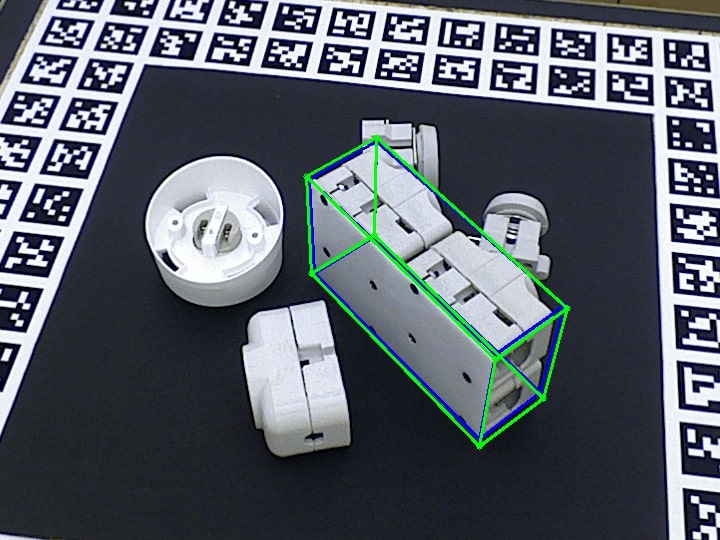} &
      \includegraphics[width=\imsize]{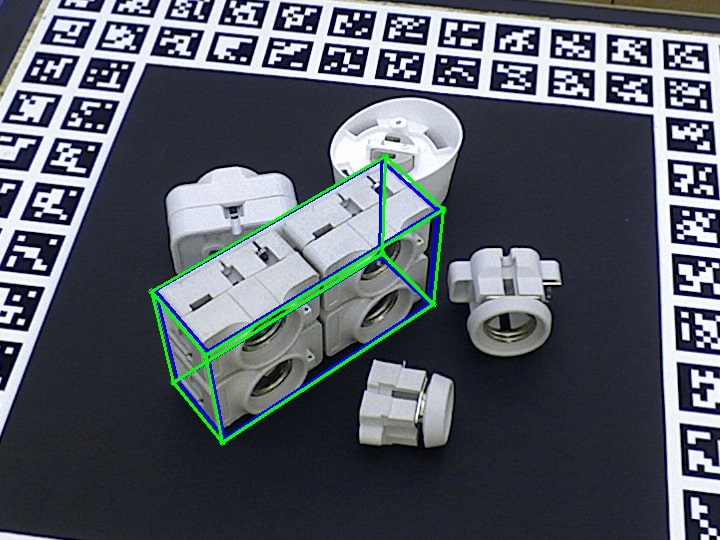} \\
    \end{tabular}
  \end{center}
  \vspace{-0.7cm}
  \caption{\label{fig:S3O8} Some qualitative results on Object \#8 in Scene \#03 of the T-LESS dataset. First
    row: 2D detection results. Second row: 3D Pose Estimation results. Green and
    blue bounding boxes are the ground truth and estimated poses respectively.}
\end{figure*}

\begin{figure*}[h!]
  \begin{center}
    \begin{tabular}{ccccc}
      \includegraphics[width=\imsize]{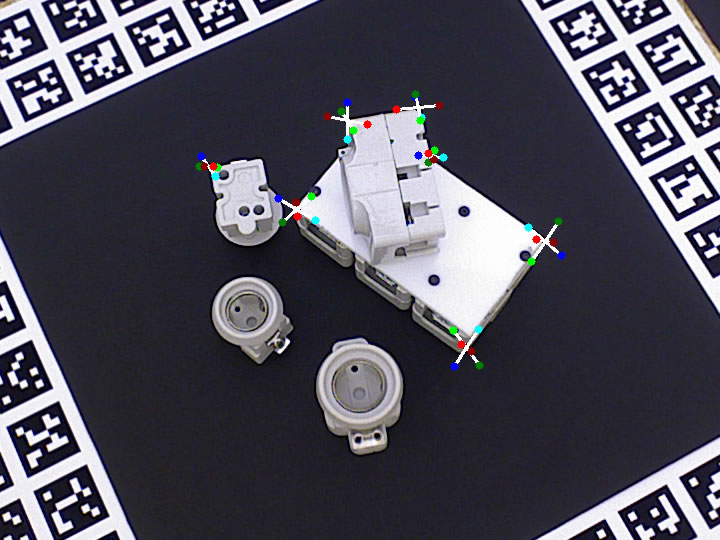} &
      \includegraphics[width=\imsize]{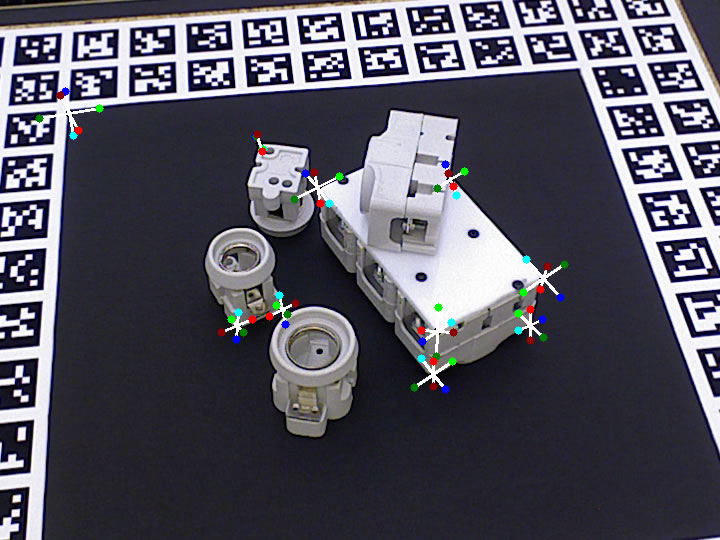} &
      \includegraphics[width=\imsize]{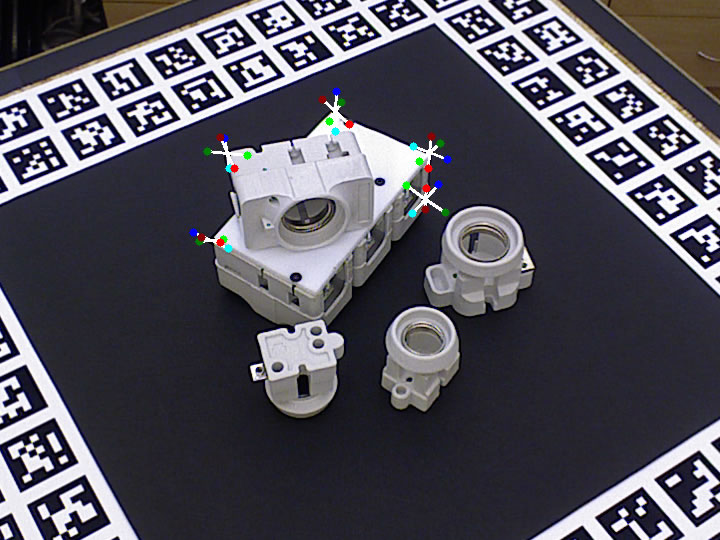} &
      \includegraphics[width=\imsize]{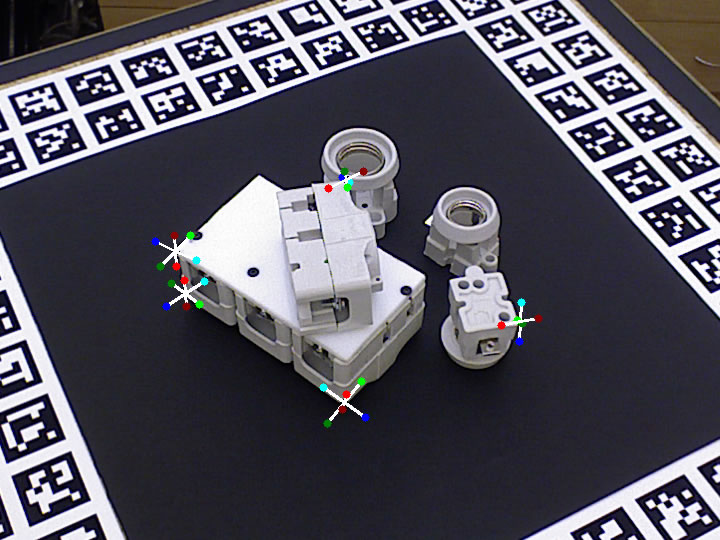} &
      \includegraphics[width=\imsize]{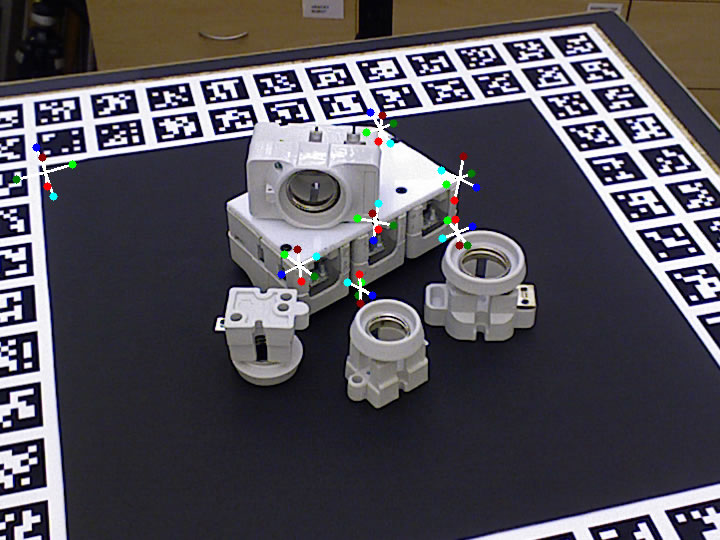} \\
      \includegraphics[width=\imsize]{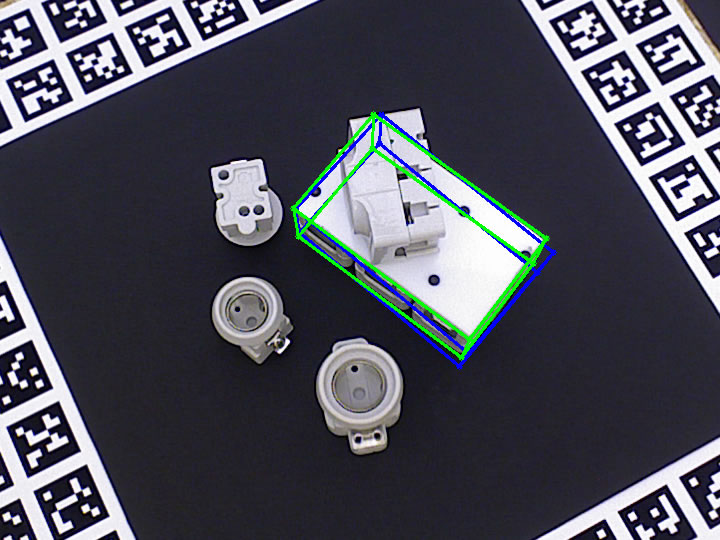} &
      \includegraphics[width=\imsize]{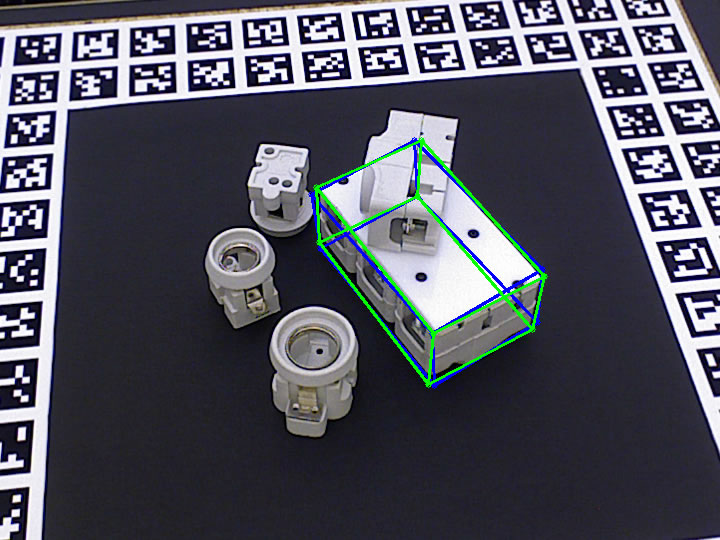} &
      \includegraphics[width=\imsize]{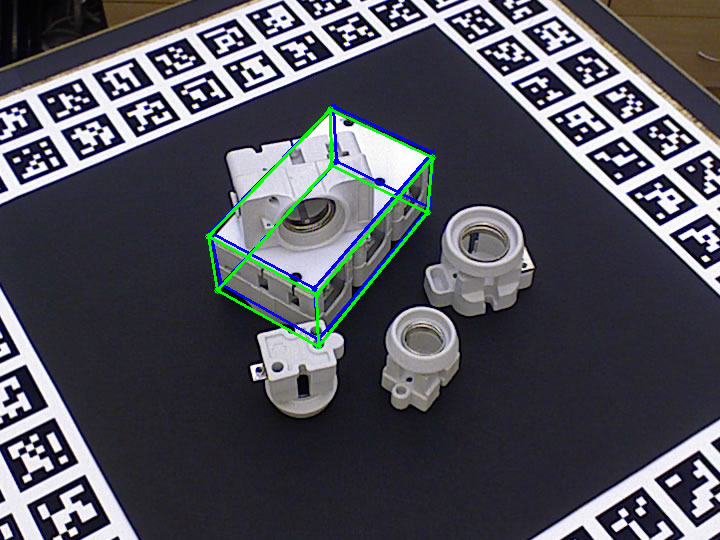} &
      \includegraphics[width=\imsize]{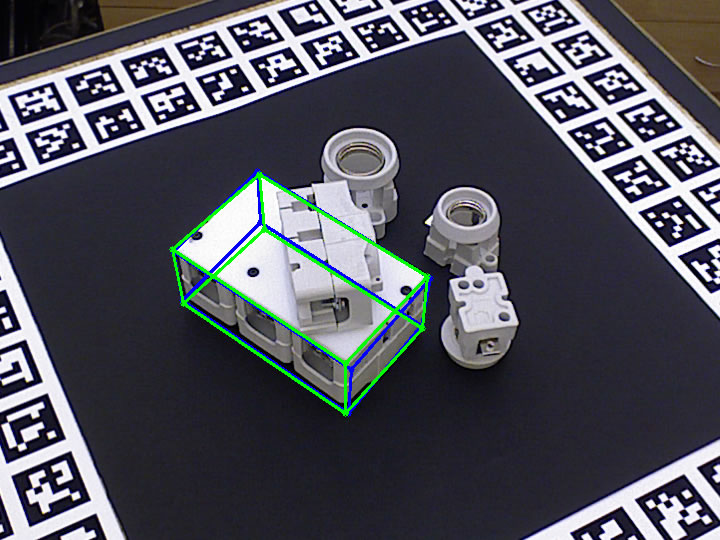} &
      \includegraphics[width=\imsize]{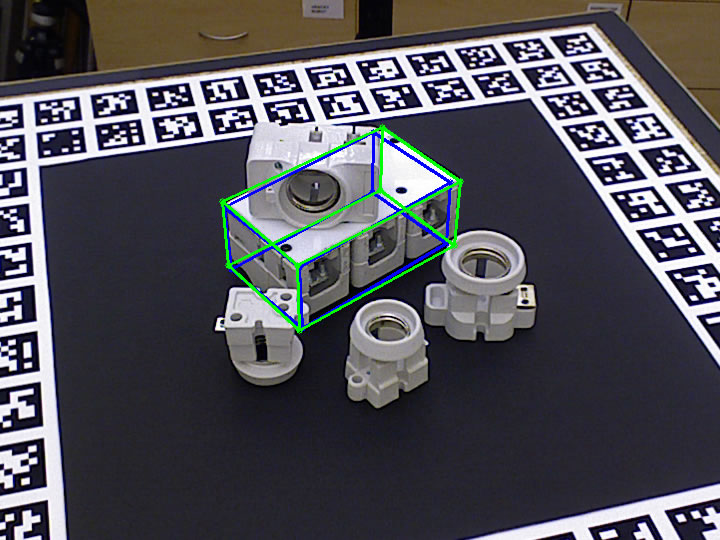} \\
    \end{tabular}
  \end{center}
  \vspace{-0.7cm}
  \caption{\label{fig:S6O7} Some qualitative results on Object \#7 in Scene \#06 of the T-LESS dataset.}
\end{figure*}

\begin{figure*}[h!]
  \begin{center}
    \begin{tabular}{ccccc}
      \includegraphics[width=\imsize]{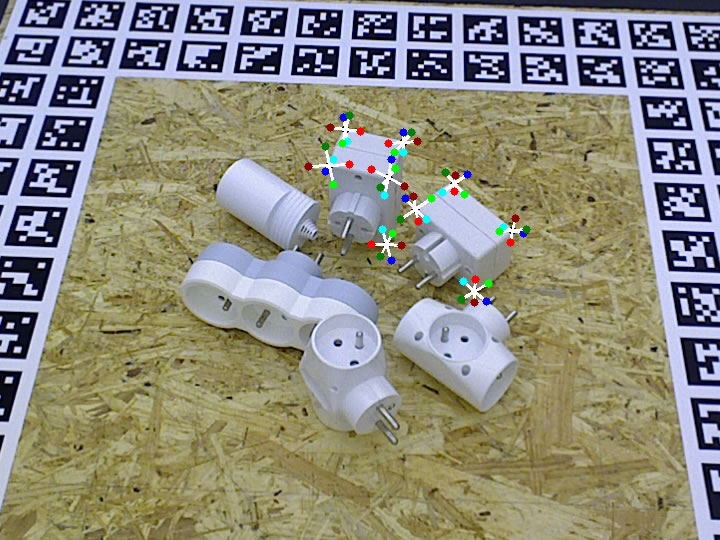} &
      \includegraphics[width=\imsize]{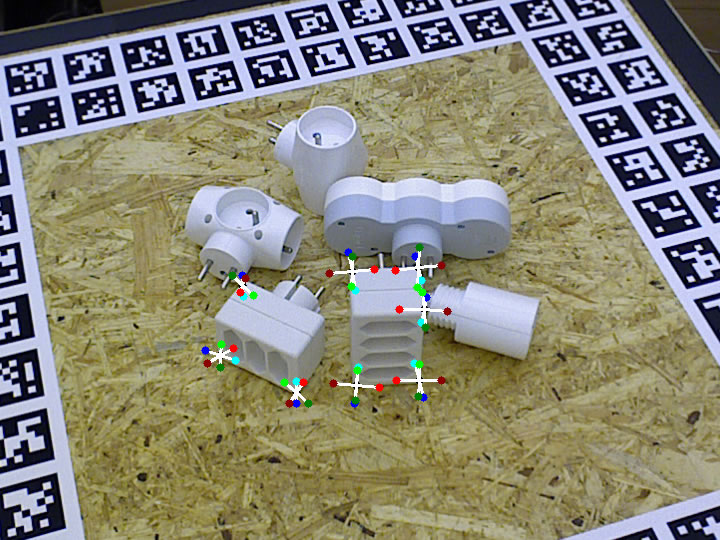} &
      \includegraphics[width=\imsize]{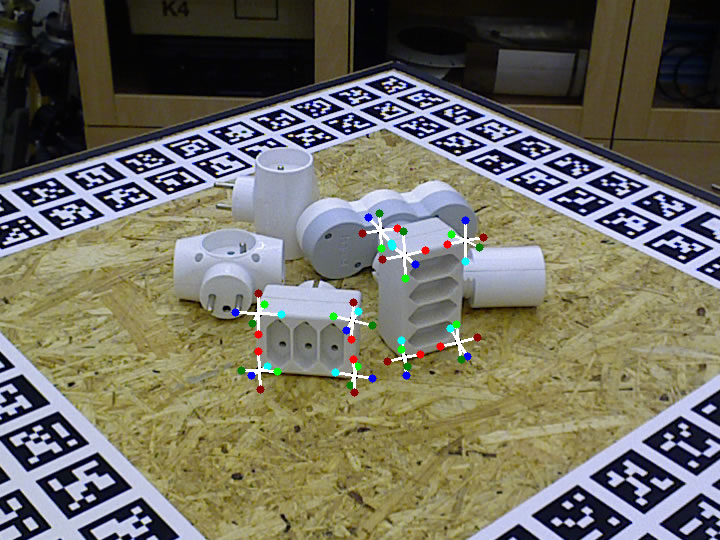} &
      \includegraphics[width=\imsize]{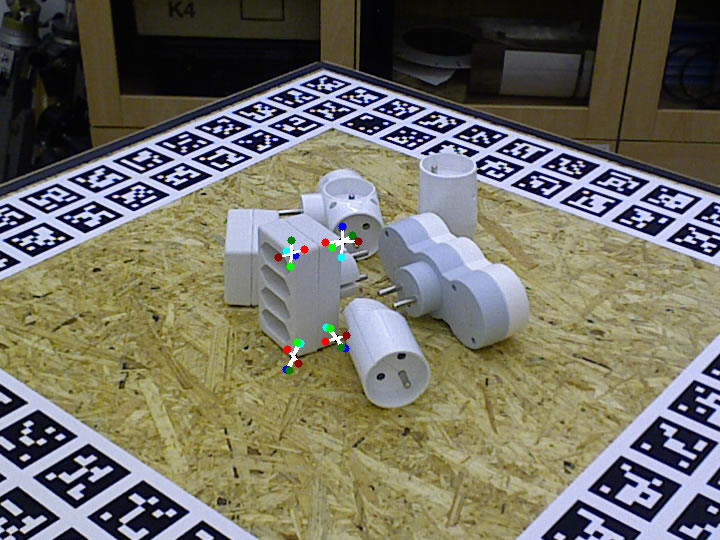} &
      \includegraphics[width=\imsize]{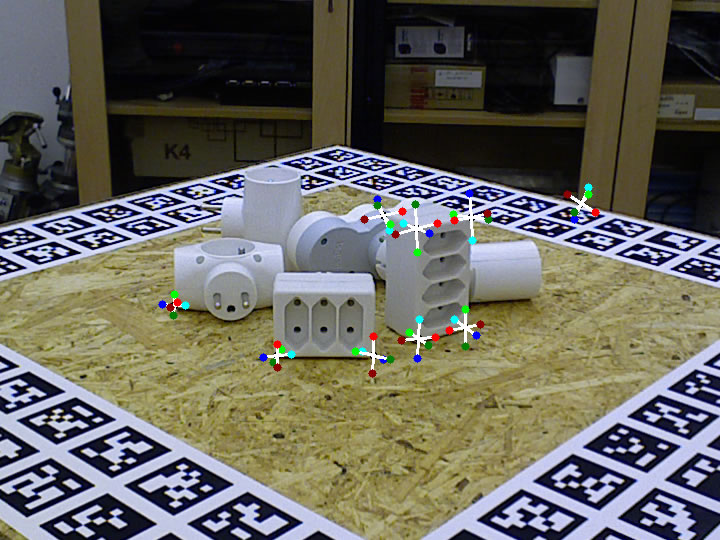} \\
      \includegraphics[width=\imsize]{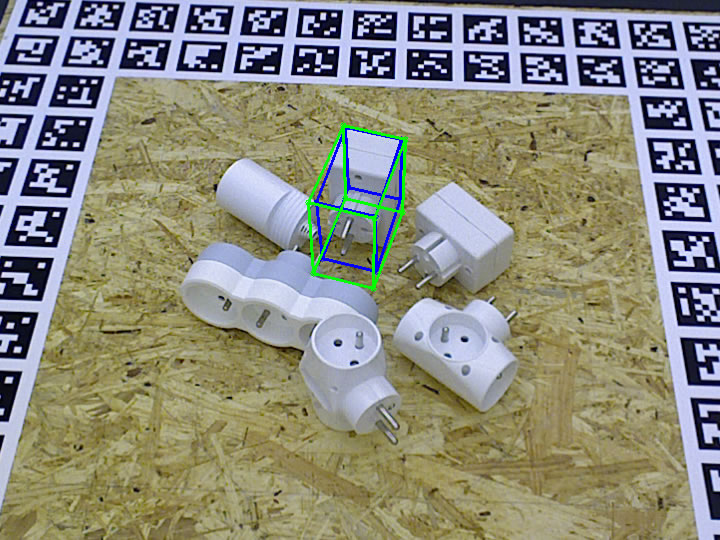} &
      \includegraphics[width=\imsize]{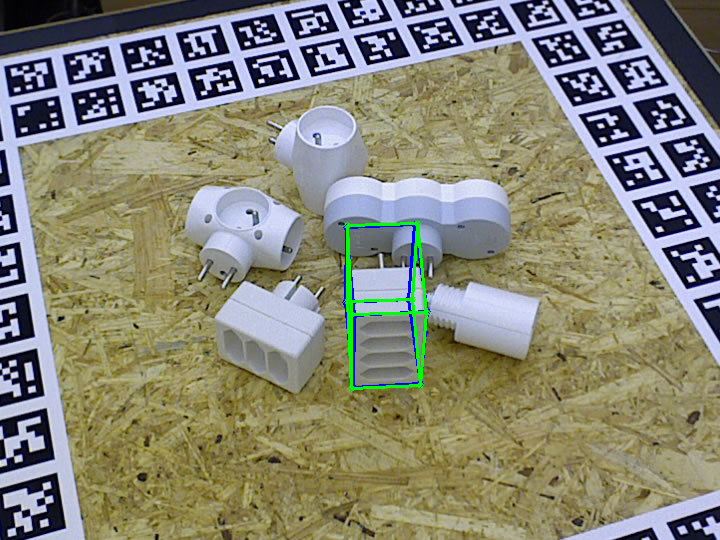} &
      \includegraphics[width=\imsize]{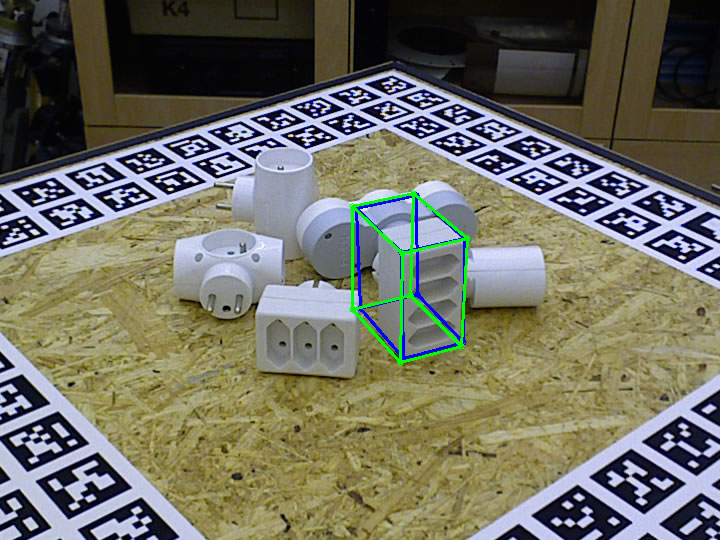} &
      \includegraphics[width=\imsize]{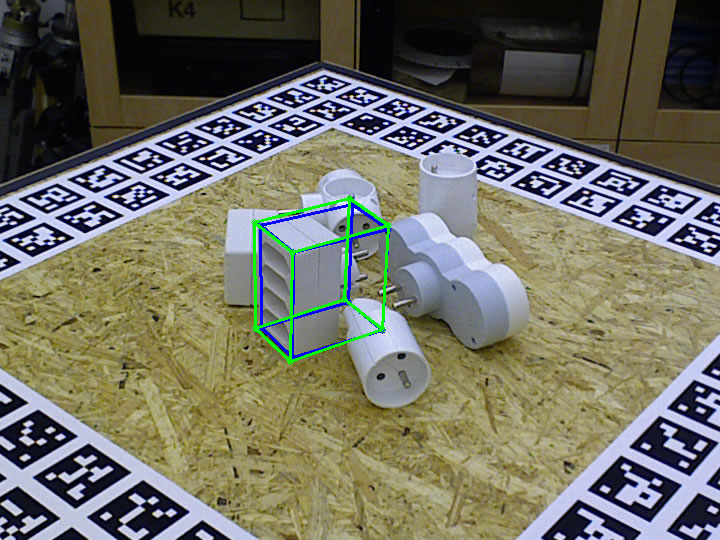} &
      \includegraphics[width=\imsize]{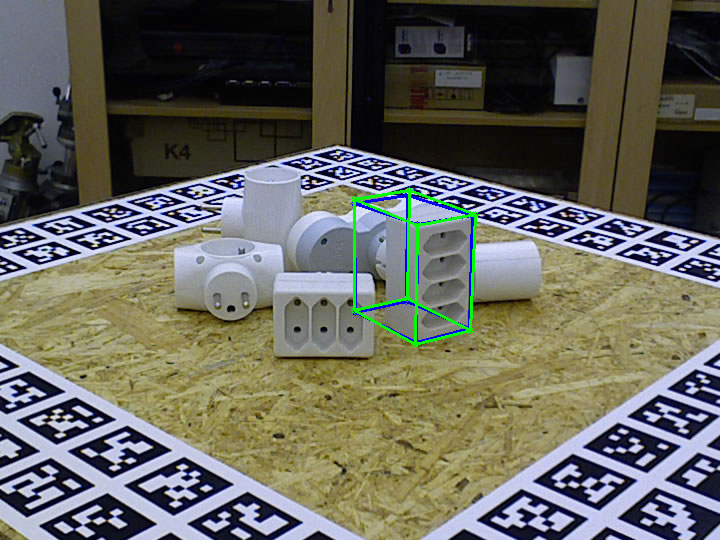} \\
    \end{tabular}
  \end{center}
  \vspace{-0.7cm}
  \caption{\label{fig:S10O20} Some qualitative results on Object \#20 in Scene \#10 of the T-LESS dataset.}
\end{figure*}

\begin{figure*}[h!]
  \begin{center}
    \begin{tabular}{ccccc}
      \includegraphics[width=\imsize]{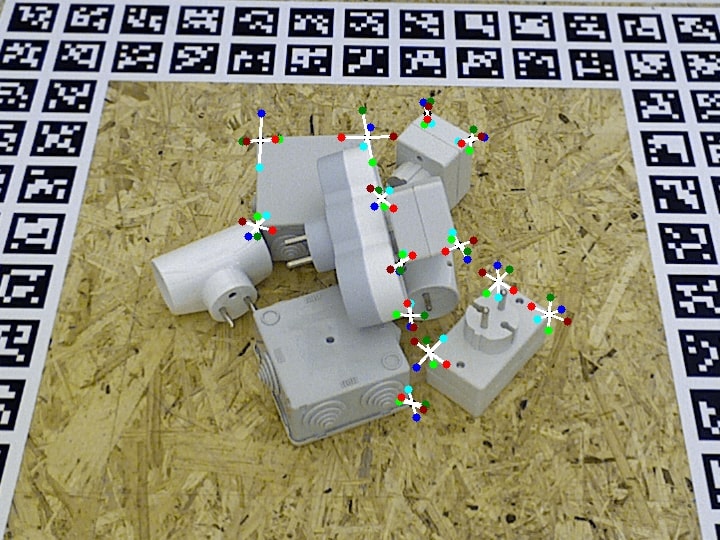} &
      \includegraphics[width=\imsize]{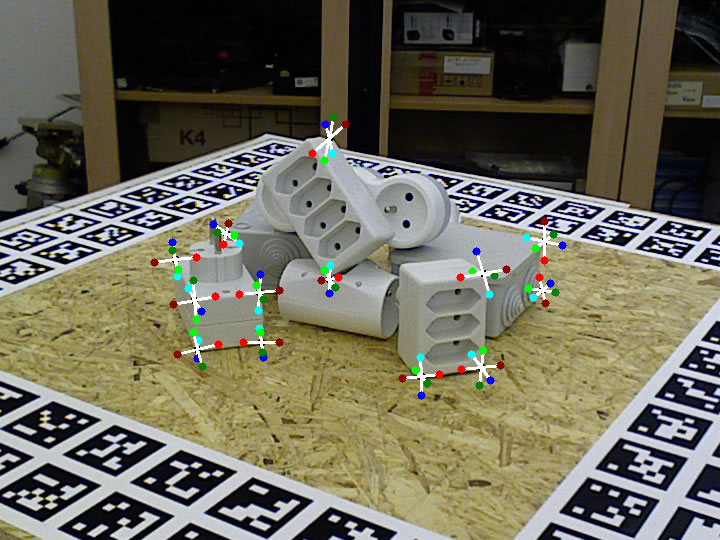} &
      \includegraphics[width=\imsize]{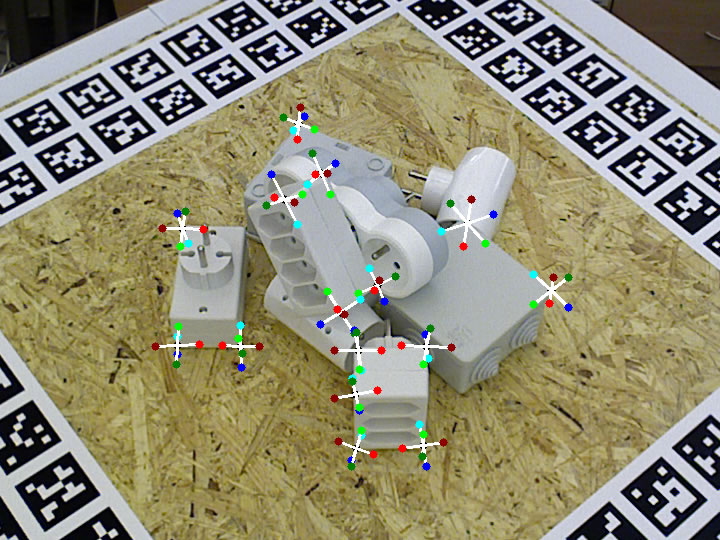} &
      \includegraphics[width=\imsize]{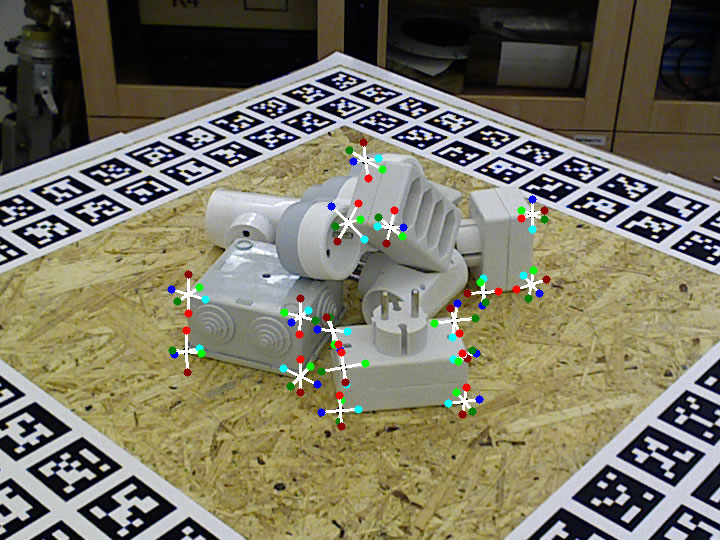} &
      \includegraphics[width=\imsize]{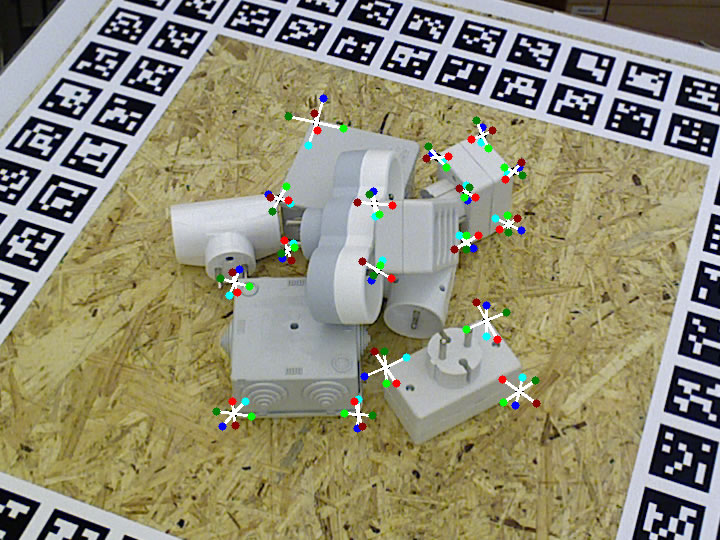} \\
      \includegraphics[width=\imsize]{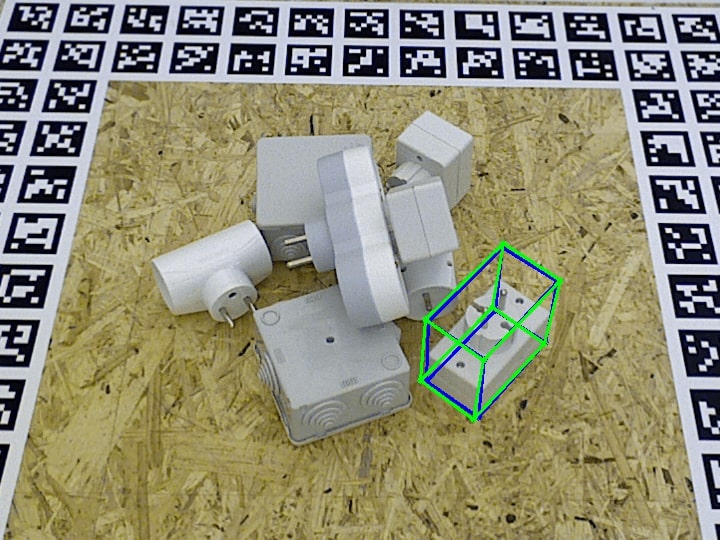} &
      \includegraphics[width=\imsize]{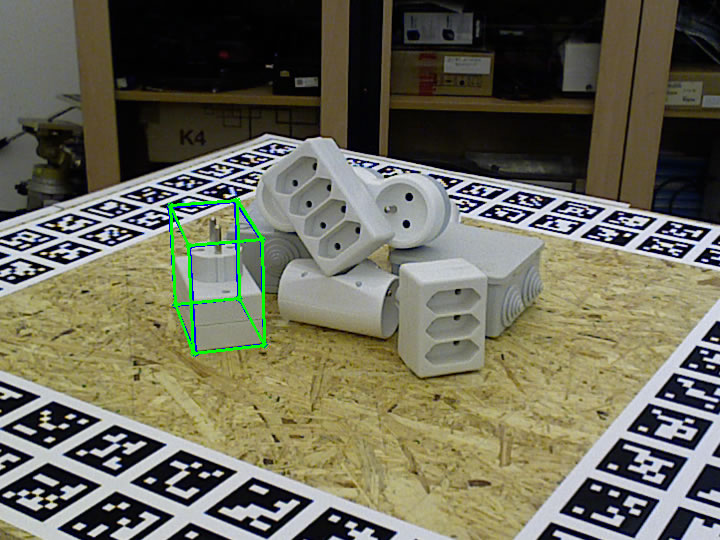} &
      \includegraphics[width=\imsize]{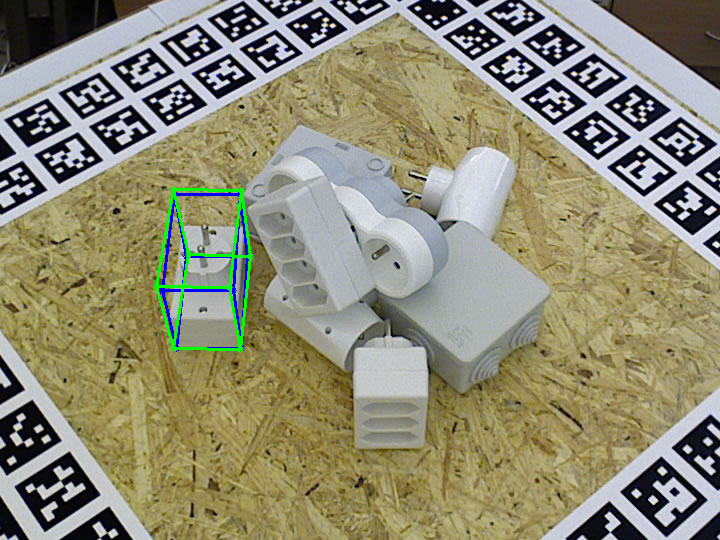} &
      \includegraphics[width=\imsize]{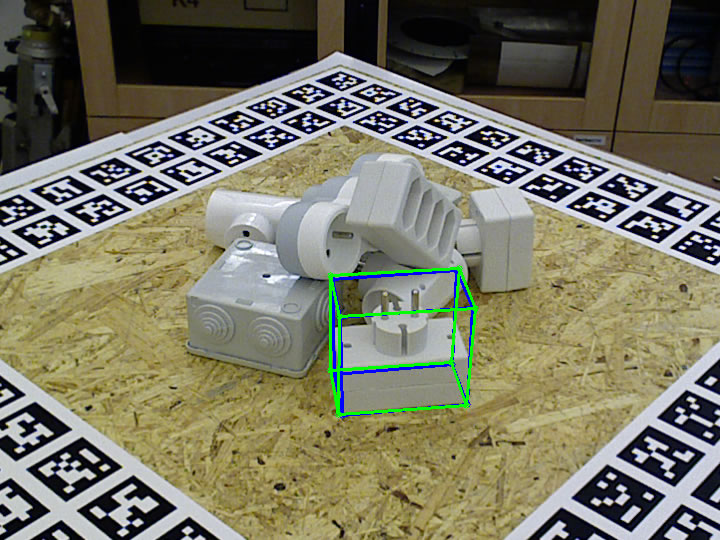} &
      \includegraphics[width=\imsize]{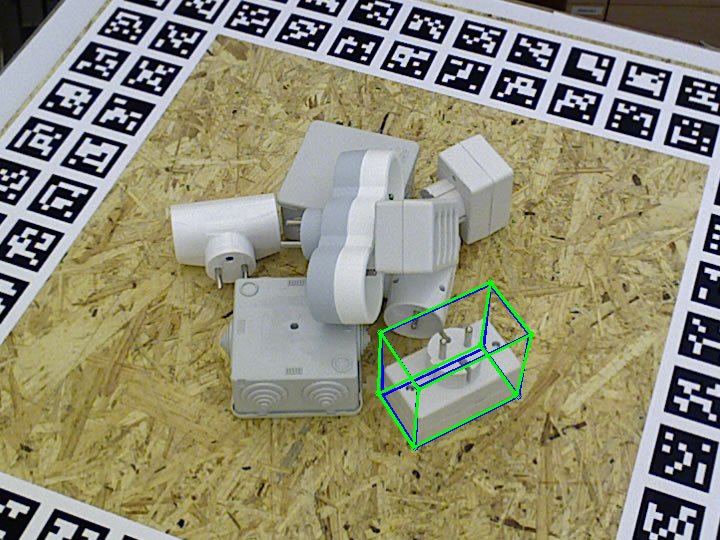} \\
    \end{tabular}
  \end{center}
  \vspace{-0.7cm}
  \caption{\label{fig:S13O20} Some qualitative results on Object \#20 in Scene \#13 of the T-LESS dataset.}
\end{figure*}

\begin{figure*}[h!]
  \begin{center}
    \begin{tabular}{ccccc}
      \includegraphics[width=\imsize]{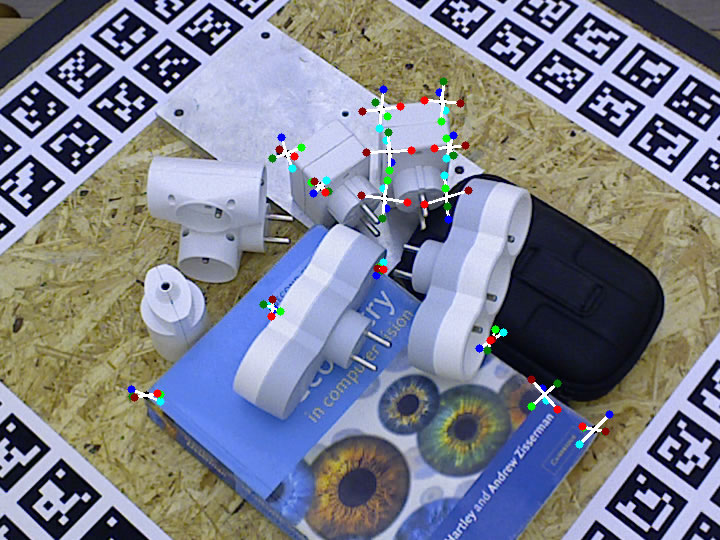} &
      \includegraphics[width=\imsize]{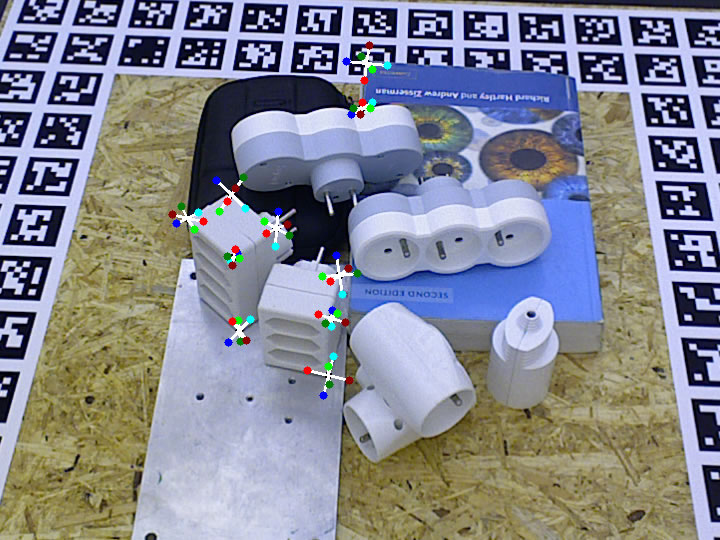} &
      \includegraphics[width=\imsize]{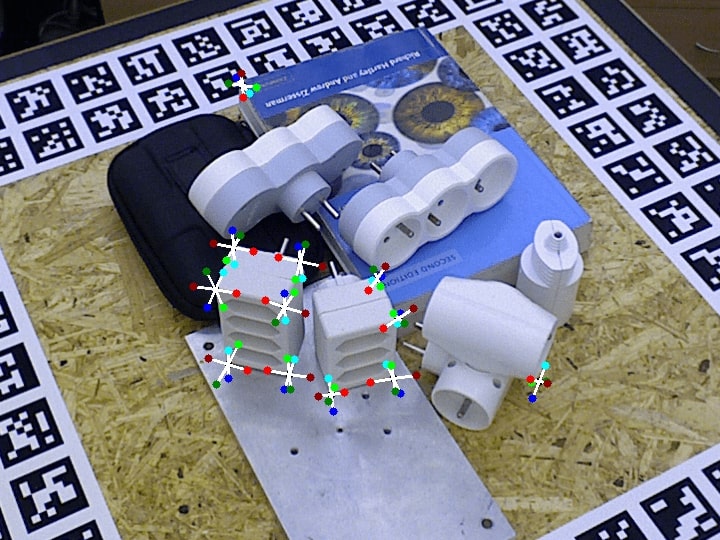} &
      \includegraphics[width=\imsize]{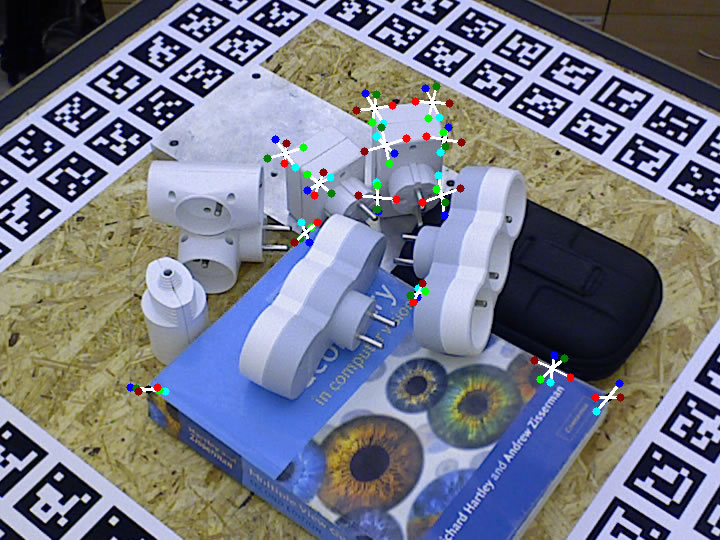} &
      \includegraphics[width=\imsize]{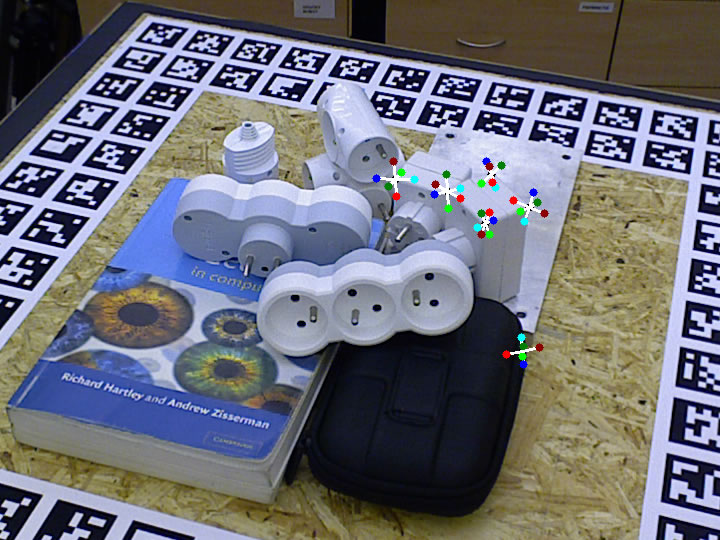} \\
      \includegraphics[width=\imsize]{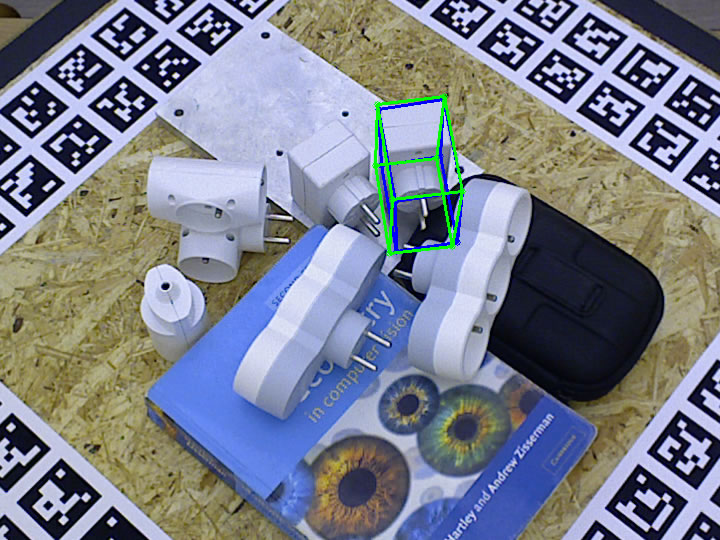} &
      \includegraphics[width=\imsize]{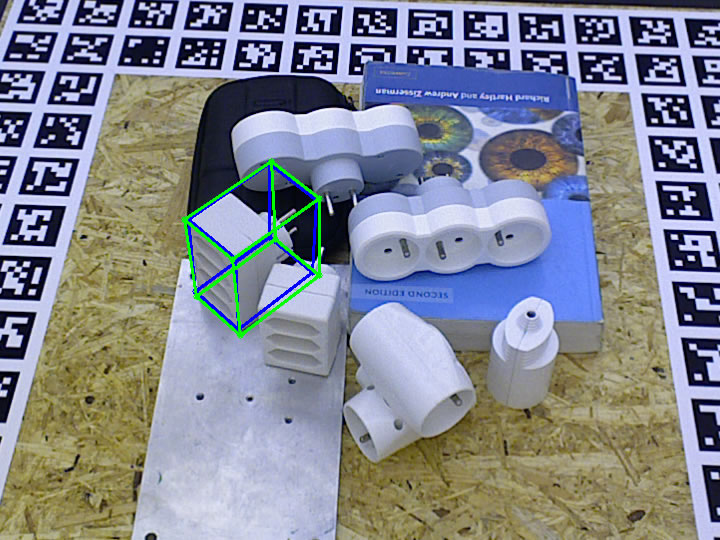} &
      \includegraphics[width=\imsize]{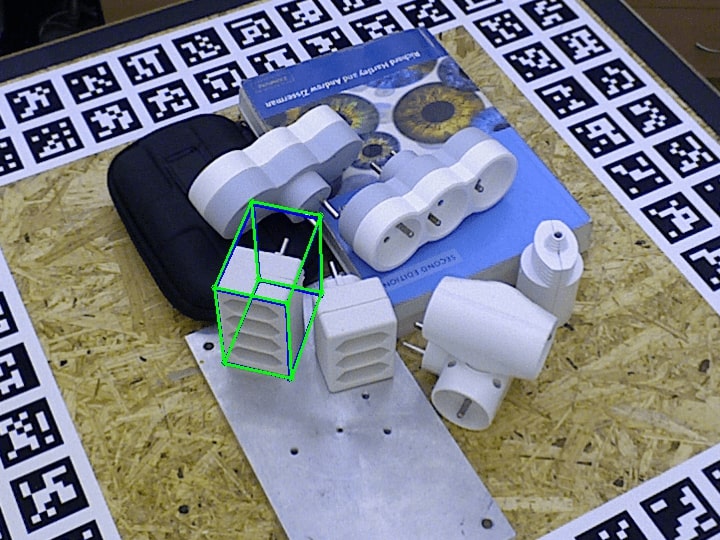} &
      \includegraphics[width=\imsize]{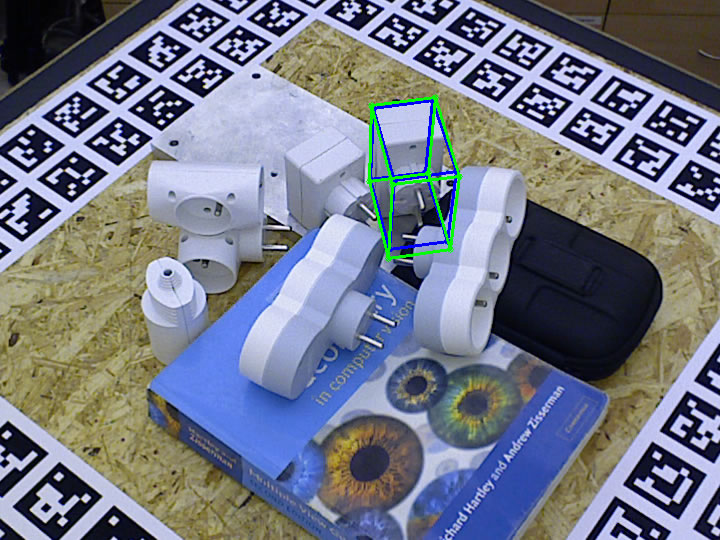} &
      \includegraphics[width=\imsize]{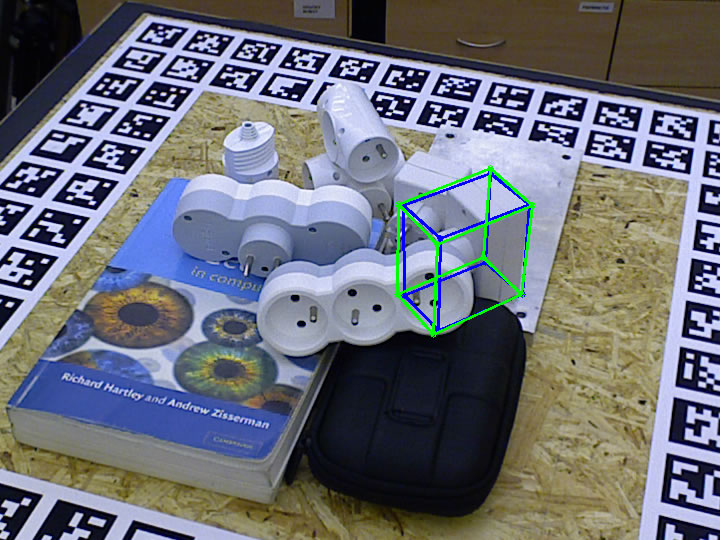} \\
    \end{tabular}
  \end{center}
  \vspace{-0.7cm}
  \caption{\label{fig:S14O20} Some qualitative results on Object \#20 in Scene \#14 of the T-LESS dataset.}
\end{figure*}

\begin{figure*}[h!]
  \begin{center}
    \begin{tabular}{ccccc}
      \includegraphics[width=\imsize]{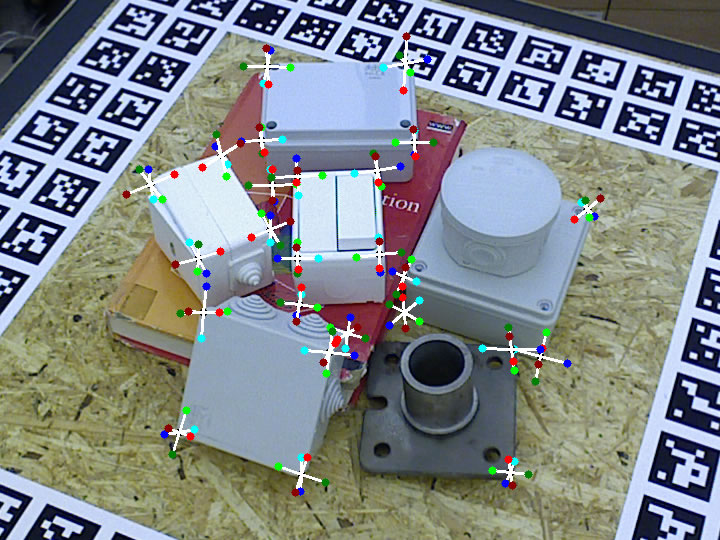} &
      \includegraphics[width=\imsize]{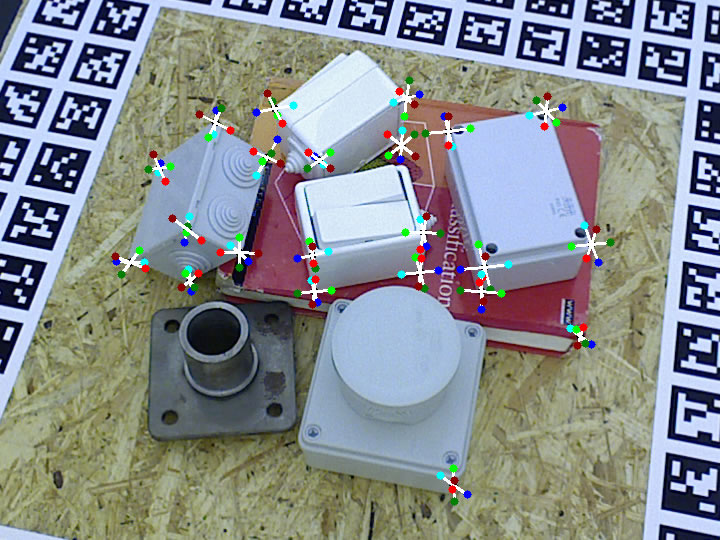} &
      \includegraphics[width=\imsize]{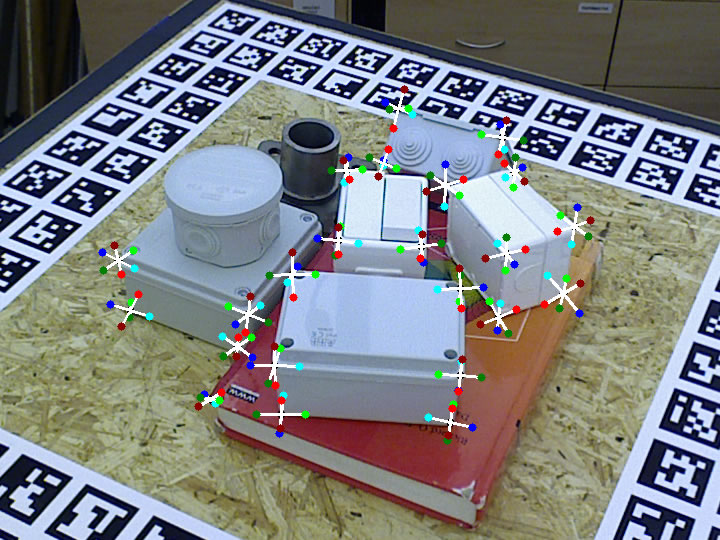} &
      \includegraphics[width=\imsize]{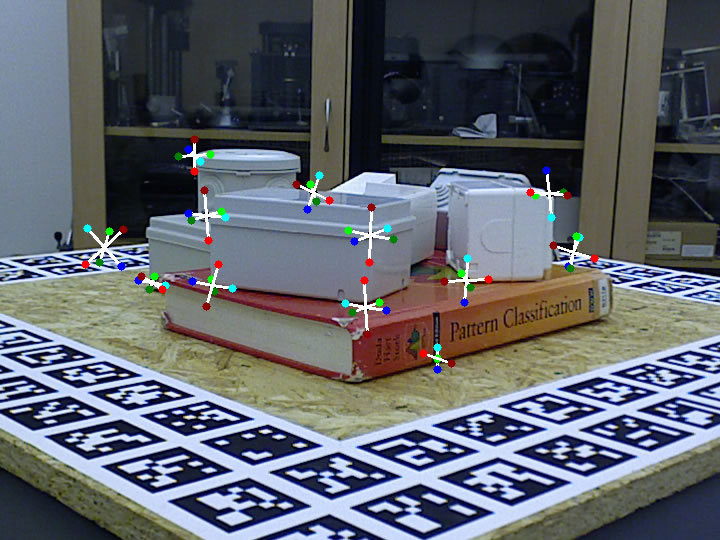} &
      \includegraphics[width=\imsize]{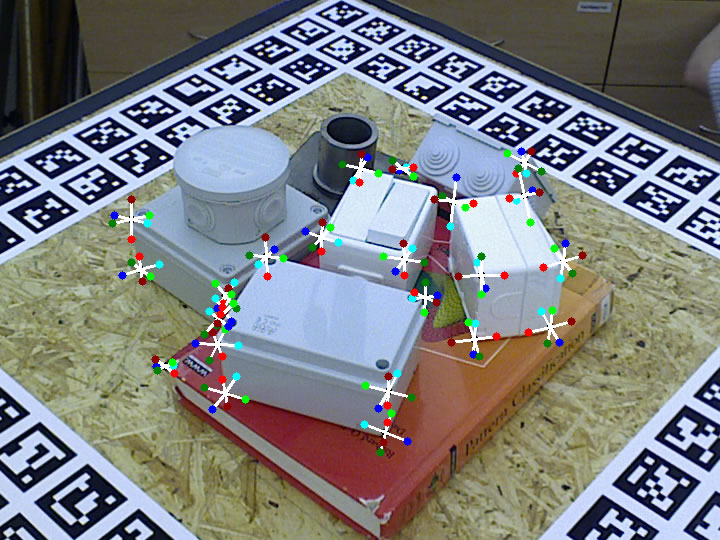} \\
      \includegraphics[width=\imsize]{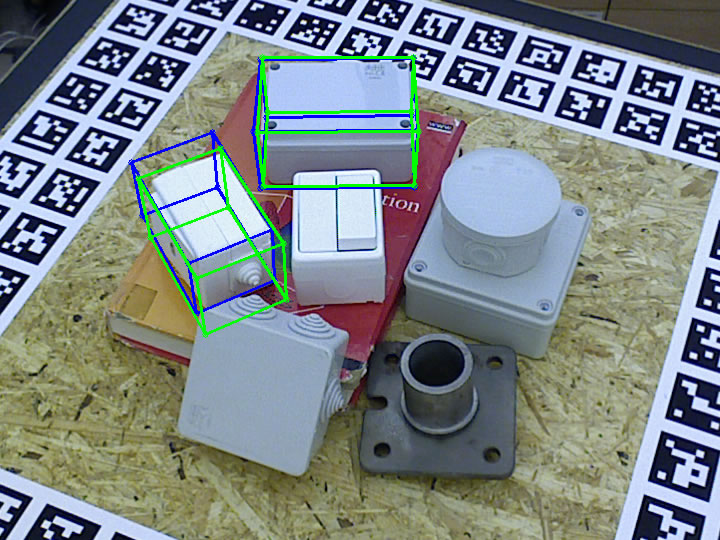} &
      \includegraphics[width=\imsize]{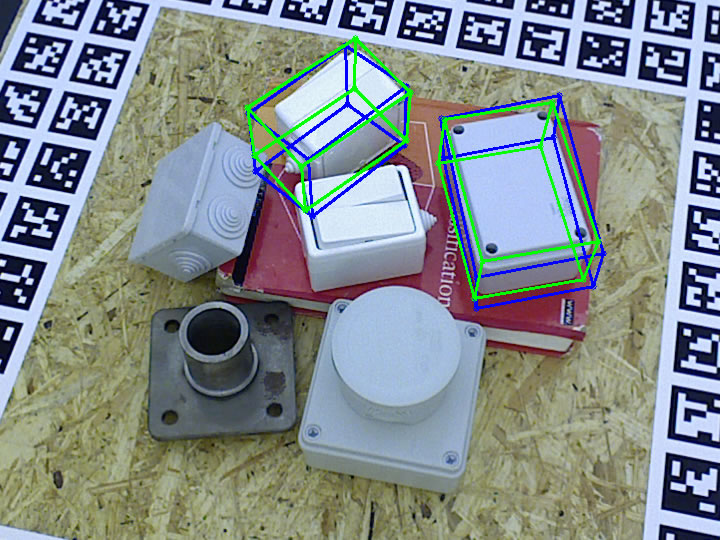} &
      \includegraphics[width=\imsize]{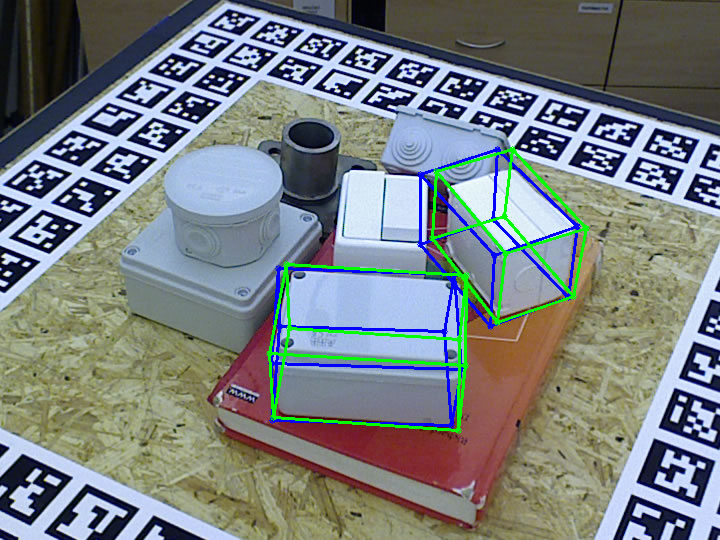} &
      \includegraphics[width=\imsize]{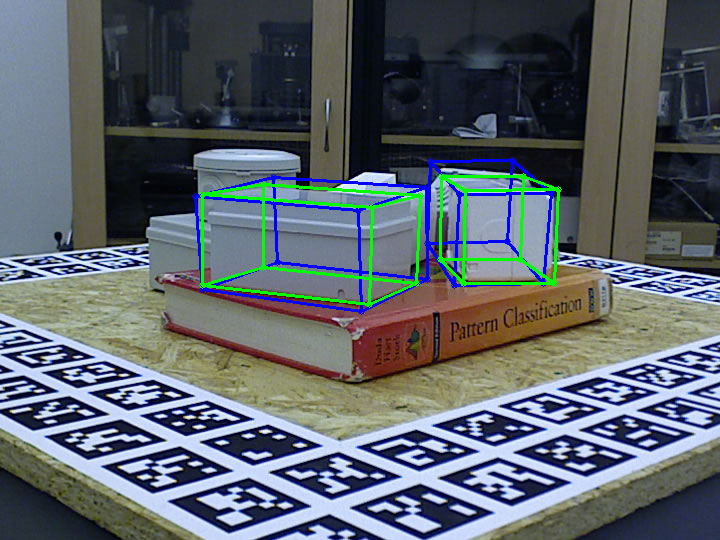} &
      \includegraphics[width=\imsize]{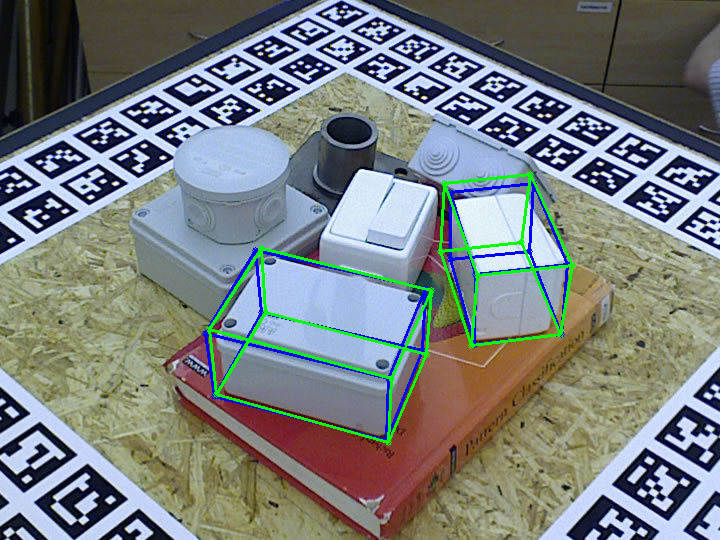} \\  
    \end{tabular}
  \end{center}
  \vspace{-0.7cm}
  \caption{\label{fig:S15O26} Some qualitative results on Object \#26 and Object \#29 in Scene \#15 of the T-LESS dataset.}
\end{figure*}

\subsubsection{Qualitative results}
To conclude the evaluation of our method, we present several qualitative results obtained on the tested scenes of the T-LESS dataset in Figs~\ref{fig:S3O8}-\ref{fig:S15O26}.  Each top row show the results of the corners detection part while each bottom row shows the estimated 3D poses. Green boxes are ground truth 3D bounding boxes while blue boxes are bounding boxes we predicted using our pose estimation pipeline.  Some scenes are very challenging. Here, the background is highly textured compared to the objects and the scenes are crowded with unwanted and close objects.  Moreover, objects seen by our network during training appear near the objects on which we wanted to test our algorithm.  Despite that, we can see that our method succeeds in estimating the pose correctly.  Moreover, Figs.~\ref{fig:S6O7} and \ref{fig:S14O20} show that detecting corners of the objects is a good direction when dealing with "crowded" scenes where partial occlusions often occur.

\subsection{Computation Times}
We implemented our method on an Intel Xeon CPU E5-2609 v4 1.70GHz desktop with a
GPU  Quadro P5000.   Our  current implementation  takes 300ms  for  the 3D  part
detection and  2s for the pose  estimation, where most  of the time is  spent in
rendering and  cross-correlation.  We believe  this part could  be significantly
optimized.

%Our implementation takes 300 ms for the 3D part detection and 150 ms for the pose estimation on an Intel %Xeon CPU E5-2609 v4 1.70GHz desktop with a GPU Quadro P5000. 
%% The  search of ambiguities during the pose estimation doesn't require too much time. Note that 60 ms of the pose estimation time are necessary to find the best pose.

%-------------------------------------------------------------------------

\section{Conclusion}
%% -*- mode: latex; mode: flyspell -*-

We  introduced a  novel approach  to  the detection  and 3D  pose estimation  of
industrial objects  in color  images that  only requires the  CAD models  of the
objects, and  \emph{no retraining}  is needed  for new  objects. We  showed that
estimating the 3D  poses of the corners  makes our method able  to solve typical
ambiguities  that raise  with industrial  objects.  A  natural extension  of our
method would  be to  consider other  types of  parts, such  as edges  or quadric
surfaces.

%-------------------------------------------------------------------------

{\small
\bibliographystyle{ieee}
\bibliography{string,vision,biblio}
}

\end{document}